\documentclass{article}

 \usepackage[dblblindworkshop, final]{neurips_2025}
\workshoptitle{Mechanistic Interpretability}

\usepackage[utf8]{inputenc} 
\usepackage[T1]{fontenc}    
\usepackage{hyperref}       
\usepackage{url}            
\usepackage{booktabs, floatrow}       
\usepackage{amsfonts}       
\usepackage{nicefrac}       
\usepackage{microtype}      
\usepackage{xcolor}         
\usepackage{graphicx}
\usepackage{caption}
\usepackage{subcaption}
\usepackage{tabularx}
\usepackage{tabularray}
\usepackage{ulem}

\definecolor{myblue}{RGB}{108, 166, 230}
\definecolor{myred}{RGB}{194, 4, 0}

\usepackage{amsmath,amsfonts,bm}

\def\eqref#1{equation~\ref{#1}}

\def\1{\bm{1}}

\def\eps{{\epsilon}}

\def\rvb{{\mathbf{b}}}

\def\rvx{{\mathbf{x}}}

\def\mH{{\bm{H}}}

\def\mS{{\bm{S}}}
\def\mT{{\bm{T}}}

\def\mW{{\bm{W}}}
\def\mX{{\bm{X}}}

\def\mZ{{\bm{Z}}}

\DeclareMathAlphabet{\mathsfit}{\encodingdefault}{\sfdefault}{m}{sl}
\SetMathAlphabet{\mathsfit}{bold}{\encodingdefault}{\sfdefault}{bx}{n}

\title{Interpretability for Time Series Transformers using \\ A Concept Bottleneck Framework}

\author{
  {\bfseries
  Angela van Sprang\textsuperscript{1},
  Erman Acar\textsuperscript{1},
  Willem Zuidema\textsuperscript{1}} \\
  \textsuperscript{1}University of Amsterdam\\
  \texttt{\{a.v.vansprang,e.acar,w.h.zuidema@\}@uva.nl}
}

\begin{document}

\maketitle

\begin{abstract}

Mechanistic interpretability focuses on \textit{reverse engineering} the internal mechanisms learned by neural networks. We extend our focus and propose to mechanistically \textit{forward engineer} using our framework based on Concept Bottleneck Models. In the context of long-term time series forecasting, we modify the training objective to encourage a model to develop representations which are similar to predefined, interpretable concepts using Centered Kernel Alignment. This steers the bottleneck components to learn the predefined concepts, while allowing other components to learn other, undefined concepts. We apply the framework to the Vanilla Transformer, Autoformer and FEDformer, and present an in-depth analysis on synthetic data and on a variety of benchmark datasets. We find that the model performance remains mostly unaffected, while the model shows much improved interpretability. Additionally, we verify the interpretation of the bottleneck components with an intervention experiment using activation patching.

\end{abstract}

\section{Introduction}
\label{Introduction}

Transformers show great success for various types of sequential data, including language \citep{devlin_bert_2018, brown_language_2020}, images \citep{dosovitskiy_image_2021, liu_swin_2021}, and speech \citep{baevski_wav2vec_2020}. Their ability to capture long-term dependencies has triggered substantial interest in applying them to time-series, which are naturally sequential, and in particular to the challenging task of long-term time series forecasting. Transformer-based architectures, indeed, often show superior performance on this task \citep{zhou_informer_2021, zhou_fedformer_2022, wu_autoformer_2021, ni_basisformer_2023, chen_pathformer_2024}, for an overview we refer to \citet{wen_transformers_2023}. 

However, due to their deep and complex architecture, transformers are difficult to interpret, which is especially important in high-stakes domains such as finance and energy demand prediction. There is a large body of work in the field of explainable AI to interpret neural networks \citep{bereska_mechanistic_2024}, or increase their interpretability, including the approach of Concept Bottleneck Models (CBMs; \citealp{koh_concept_2020}). This approach relies on the idea of constraining the model such that it first predicts human-interpretable concepts, and then uses only these concepts to make the final prediction. 
CBMs and their variants have become popular in various fields, especially in computer vision, but are so far unexplored in the context of time series forecasting.

\begin{figure}
    \centering
    \includegraphics[width=0.9\linewidth]{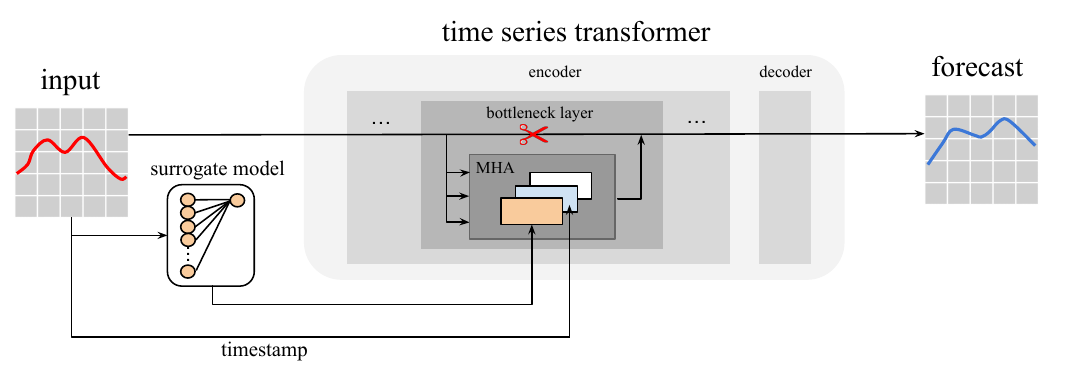}
    \caption{Overview of the concept bottleneck framework. The bottleneck is one encoder layer which is trained to be similar to pre-defined, interpretable concepts. The residual stream around the bottleneck is removed, such that all information passes through the bottleneck.
    }
    \label{fig:overview}
\end{figure}

In this paper, we propose a training framework to make any time series transformer into a Concept Bottleneck Model using time-series specific, yet domain-agnostic concepts, as shown in Figure \ref{fig:overview}. A key aspect of our training framework is to leave the model's architecture intact, while encouraging the learned representations to be similar - but not identical - to the interpretable concepts. 
We measure similarity with Centered Kernel Alignment (CKA; \citealp{kornblith_similarity_2019}) and include it in the loss function. 
The first concept is a simple, linear surrogate model and the second is time information (e.g. hour-of-day). 
Note that we propose a \textit{global} interpretability method, which improves identifying and localizing high-level concepts in the model's internal mechanisms, and is not comparable to local post-hoc interpretability methods such as SHAP, LIME, or attention-based visualizations which explain individual predictions.



We apply our concept bottleneck framework to three types of models: Vanilla Transformer {\citep{vaswani_attention_2017}}, Autoformer \citep{wu_autoformer_2021} and FEDformer \citep{zhou_fedformer_2022}. Across extensive experiments on seven datasets, we show that our setup results in models that are more interpretable while the overall performance remains largely unaffected -- in many cases surpassing results from the original Autoformer paper. 
Furthermore, we explicitly test the faithfulness of the obtained interpretations with an intervention study using activation patching.

Our contributions are summarized as follows:
\begin{enumerate}
    \item We propose a novel training framework to increase the interpretability of transformers for time series.
    \item We demonstrate the feasibility of applying this framework to time-series transformers by conducting extensive experiments on {three} types of transformers and seven datasets, {and identify interpretable concepts in each of these transformers}.
    \item We assess the faithfulness of the interpretability analysis by performing an activation patching experiment, and obtain evidence that the identified components (in the concept bottleneck) indeed have the hypothesized unique and causal role in the predictions of the target model.
\end{enumerate}

\section{Background and Related Work}
This paper combines and builds upon foundational works from different fields, including CBMs, knowledge transfer with CKA and time series transformers. CBMs have been applied to time series before \citep{ferfoglia_ecats_2024}, but not with the same interpretable concepts. Likewise, the similarity index CKA has been used before to transfer knowledge between models \citep{tian_continuous_2023}, yet, to the best of our knowledge, it has not been used to construct a CBM. This makes our work a unique contribution at the intersection of (mechanistic) interpretability, concept learning, and time series forecasting.

\subsection{Concept Bottleneck Models}
Concept Bottleneck Models (CBMs; \citealp{koh_concept_2020}) have emerged as promising interpretable models \citep{poeta_concept-based_2023}. The concept bottlenecks constrain the model to first predict interpretable concepts, and then use only these concepts in the final downstream task. They are shown to be useful in multiple applications, such as model debugging and human intervention. The bottleneck allows for explaining which information the model is using and when it makes an error due to incorrect concept predictions. 

One of the shortcomings of standard CBMs is that concept annotations are needed during training to learn the bottleneck, and concept labels do not necessarily contain all information needed to accurately perform the downstream task, and can therefore decrease the task accuracy \citep{mahinpei_promises_2021}. Therefore, \citet{zarlenga_concept_2022} propose Concept Embedding Models, where concepts are represented as vectors, such that richer and more meaningful concept semantics can be captured.

CBMs and their variants are usually applied to the field of computer vision, and less frequently to natural language \citep{tan_interpreting_2024}, graphs \citep{barbiero_interpretable_2023} or tabular data \citep{zarlenga_concept_2022}. In principle, the methodology can be applied to time series as well, but defining high-level, meaningful concepts is challenging. \citet{ferfoglia_ecats_2024} use Signal Temporal Logic (STL) formulas as concept embeddings for time series to convert them into natural language, and use these concepts as bottleneck for anomaly detection.


\subsection{Knowledge Transfer with Centered Kernel Alignment}
Inspired by neuroscience, CKA measures the similarity between different representations from neural networks \citep{kornblith_similarity_2019}. By factoring out differences in scaling or orthogonal transformations, CKA captures intuitive notions of similarity between representations. 
To obtain the score, firstly, the similarity between every pair of examples in each representation separately is measured using a pre-defined kernel, and then the obtained similarity structures are compared. We use a linear kernel, which makes the CKA score defined as follows for representations $X$ and $Y$:
\begin{equation}
    \operatorname{CKA}(X, Y)=\frac{\left\|Y^{\top} X\right\|_F^2}{\left\|X^{\top} X\right\|_F\left\|Y^{\top} Y\right\|_F}.
\end{equation}

The CKA score can be used to transfer knowledge between different models when included in the loss function \citep{tian_continuous_2023}. 
In this work, the authors study knowledge distillation between a teacher and student model, and incorporate CKA into the loss function to transfer feature representation knowledge from the pretrained model to the incremental learning model \citep{parisi_continual_2019}. 

\subsection{Time Series Transformers} \label{sect: autoformer}
Time series transformers for long-term time series forecasting, such as the Autoformer and FEDformer, obtain two types of input: (1) \emph{data values} $\mX \in \mathbb{R}^{I \times d}$, and (2) \emph{timestamps} $\mT \in \mathbb{R}^{I \times 4}$. More specifically, they can be regarded as a function $f: \mathbb{R}^{I \times d} \times \mathbb{R}^{I \times 4} \times \mathbb{R}^{O \times 4} \rightarrow \mathbb{R}^{O \times d}$, where $I$ is the number of input time steps, $O$ is the number of future time steps, and $d$ is the number of variables in the time series. The additional four dimensions of timestamps $\mT$ represent four time features, namely \emph{hour-of-day, day-of-week, day-of-month}, and \emph{day-of-year}. The future timestamps are also provided, for which the model should forecast the future data values. Note that we explicitly introduce a notation for the timestamps to later define the CKA scores and the intervention.


\color{black}

\section{Method} \label{sect: method}

\begin{figure}[]
\captionsetup{}
    \centering
    \includegraphics[width=.8\linewidth]{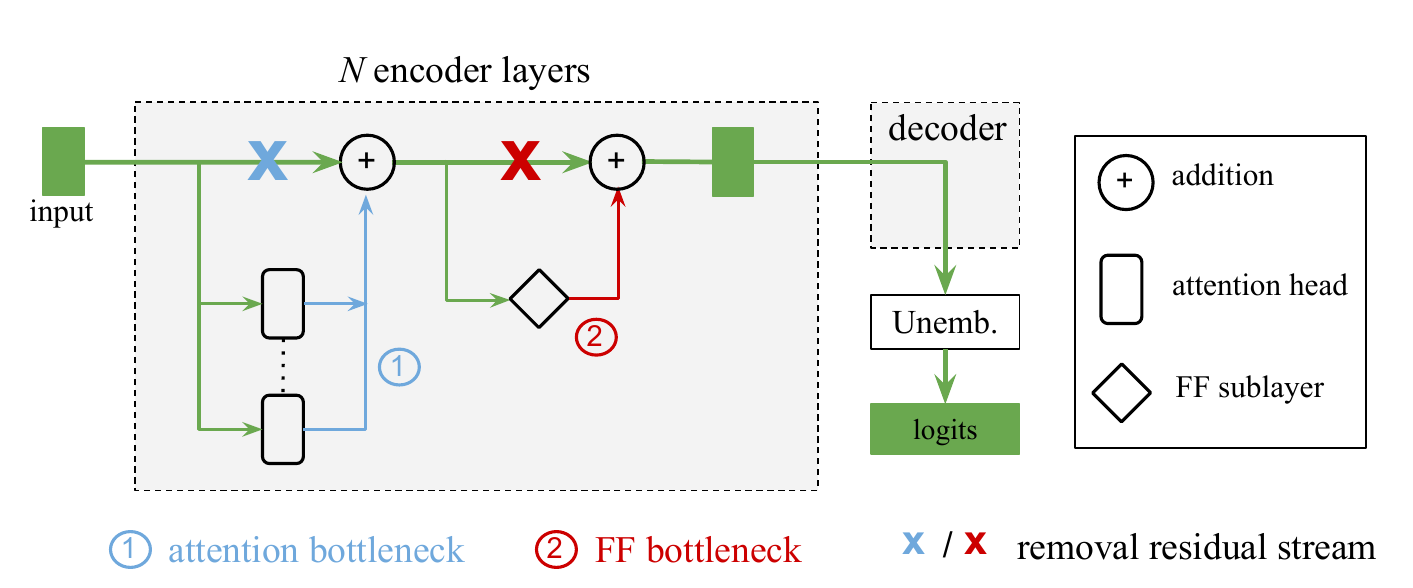}
    \caption{Architecture of a transformer with a concept bottleneck in the attention mechanism (\textcolor{myblue}{blue}) or the FF network (\textcolor{myred}{red}). Note that the residual connection is removed at the location of the bottleneck (and the residual stream thus interrupted). 
    Visualisation inspired by \citealp{rai_practical_2024}.
    }
    \label{fig:overview bottleneck}
\end{figure}

We propose a training framework to make any transformer model interpretable by including a bottleneck based on knowledge transfer with CKA  \citep{kornblith_similarity_2019}, as shown in Figure \ref{fig:overview bottleneck}. The main idea is that we assign one of the encoder layers to be the \emph{concept bottleneck}; representations in the bottleneck are subject to a soft constraint of being as similar as possible to predefined interpretable concepts. To this end, 
we calculate CKA scores with the interpretable concepts, and include these scores in the loss function. 

\subsection{Loss Function} \label{sec:loss}
The loss function should encourage the model to represent the interpretable concepts in the bottleneck layer. Therefore, we add a term $\mathcal{L}_{CKA}$ based on the CKA scores of the bottleneck and the interpretable concepts (Eq. \ref{eq: 2}). In particular, low similarity between the bottleneck and the interpretable concepts results in a higher value for $\mathcal{L}_{CKA}$. The total loss function $\mathcal{L}_{Total}$ (Eq. \ref{eq: 1}), then, is a weighted average of the Mean Squared Error (MSE) loss $\mathcal{L}_{MSE}$ and the CKA loss $\mathcal{L}_{CKA}$: 
\begin{align}
    \mathcal{L}_{Total} &= (1 - \alpha) \ \mathcal{L}_{MSE} + \alpha \ \mathcal{L}_{CKA}, \label{eq: 1} \\
    \mathcal{L}_{CKA} &= 1 - \frac{1}{c} \sum_{i=1}^{c} CKA_i, \label{eq: 2}
\end{align}
where $\alpha$ is a hyperparameter, $c$ is the number of concepts, and $CKA_i \in [0,1]$ is the CKA score (using a linear kernel) between the model's representation and concept $i$ (see Section \ref{sect: bottleneck}). 

\subsection{Interpretable Concepts in the Bottleneck} \label{sect: bottleneck}

In this section, we describe how to calculate the CKA score to measure the presence of a concept. We refer to Appendix \ref{app: formalization} for a more detailed description of the concept bottleneck framework. 

\textbf{Location bottleneck.}
We assign one encoder layer to be the bottleneck layer, because the encoder focuses on modelling seasonal information. Within the bottleneck layer, the latent representations can be taken from two different types of blocks: the attention block ($\tau=Att$) and the feed-forward block ($\tau=FF$). These two options are illustrated in Figure \ref{fig:overview bottleneck}. We assign $c$ interpretable concepts over the latent representations, with the goal of teaching the corresponding model component to represent the pre-defined interpretable concepts.


Since the attention block is multi-headed, different heads naturally form the components of the attention bottleneck. Moreover, the components need to be divided between the heads, which would be convenient when the number of heads is a multiple of the total number of concepts to maintain a uniform concept per head ratio. For the feed-forward bottleneck, we define the components to be slices from its output, such that stacking the components results in the original output. 


\textbf{Interpretable concepts.} For the real-world time series, we use two domain-agnostic interpretable concepts which can be used for forecasting, namely: (1) a simple, human-interpretable surrogate forecasting model, (2) the input timestamps recorded with the time series. Note that each model token (time step) should map to each concept to calculate the CKA score.
\begin{enumerate}
    \item We use a simple autoregressive model (AR) as a surrogate model, which predicts the next future value as a linear combination of its past values. This model is transparent, and the attribution of each input feature to the output can be simply interpreted by its weight. This concept can also be regarded as a baseline for the forecasting performance. The model is fit to the same training data as the transformer (with its order being the length of the input). We use its activations to calculate the CKA score.
    \item We use the hour-of-day feature from the timestamps $\mT$ as interpretable time concept, denoted by $\mT_{hourofday}$. This provides the bottleneck with a simplified notion of time.
\end{enumerate}



\textbf{Removal of residual connection.} Any transformer layer contains residual connections around the attention and feed-forward blocks. To ensure that all information passes through the bottleneck, we remove the residual connection around the bottleneck, potentially at the cost of a loss in performance. Otherwise, any concept, including the interpretable concepts, can be passed through the residual connection and compromise the bottleneck. 

In the scenario that the number of components is equal to the number of interpretable concepts ($c=2$), the construction of the bottleneck limits learning domain-specific features from the data, other than the interpretable concepts. Therefore, we perform experiments where we allow an extra component in the bottleneck to not learn any pre-defined concept ($c=3$). In other words, the extra component serves as a \emph{side-channel} or \emph{free component}, on which no CKA loss is calculated. 
The free component may partly restore the information lost by removing the residual connection, but with the advantage that we can monitor which information goes through it, and even visualize it (as in Section \ref{sec:compvisualizations}).

\subsection{Implementation details.} \label{sect: implementation} In our experiments, we use transformer models with three encoder layers, of which the bottleneck layer is the second layer. Similar to the original Autoformer paper, we use one decoder layer, employ the Adam optimizer \citep{kingma_adam_2017} with an initial learning rate of $10^{-4}$, and use a batch size of 32. The training process is early stopped within 25 epochs. All experiments are repeated five times on different seeds, using hyperparameter $\alpha = 0.3$. Each model is trained on 1 Nvidia GeForce GTX 1080 Ti with 30 GB for approximately 30 minutes.

\section{Experiments}

We evaluate our framework on three models and seven datasets, including synthetic and real-world data. The six real-world benchmarks consider the domains of energy, traffic, economics, weather, and disease, similar to \citet{wu_autoformer_2021}. These datasets are multivariate, and the task is to predict the future values of all variates. For example, the electricity dataset consists of hourly measurements of the electricity consumption of 321 customers from 2012 to 2014. For more information on the datasets, we refer the reader to Appendix \ref{app: datasets}. We apply the experiments to the Vanilla Transformer, Autoformer and FEDformer. First, we train a simple AR model on the same data, so that its outputs can be used to align the representations of the bottleneck. Then, we train the transformers with and without bottleneck, using different configurations for the bottleneck. 

\subsection{Synthetic Data}
To show the general applicability of the bottleneck framework, we first train an Autoformer on a synthetic time series. In particular, we generate the dataset as the sum of different sines using the function $f_{Total}$ with time $t$ as follows:
\begin{equation*}
    f_{Total}(t) = f_1(t) + f_2(t) + f_3(t),
\end{equation*}
where:
\begin{align*}
    f_{1}(t) &= \sin(2 \pi t), \\
    f_{2}(t) &= \frac{1}{2} \sin(4 \pi t + \frac{\pi}{4}), \\
    f_{3}(t) &= \frac{1}{4} \sin(6 \pi t + \frac{\pi}{2}) + \eps_t.
\end{align*}
Note that all functions $f_1, f_2$ and $f_3$ follow a periodic structure, and $f_3$ contains random noise $\eps$ from a normal distribution with standard deviation of 0.2.

\begin{figure}[h!]
     \centering
     \begin{subfigure}[b]{0.49\textwidth}
         \centering
         \includegraphics[width=\textwidth]{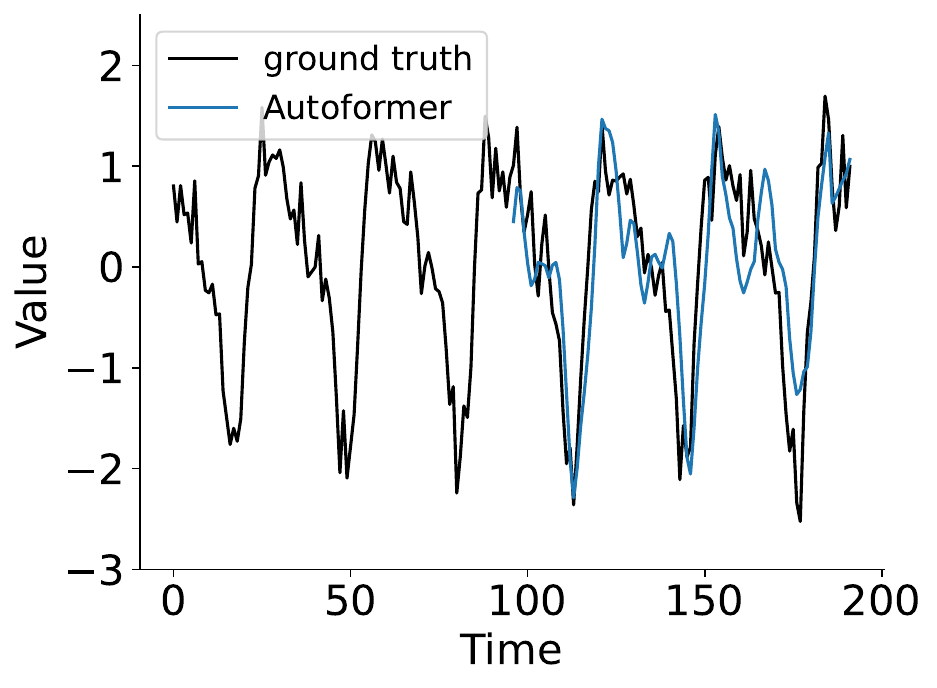}
         \caption{Forecast}
     \end{subfigure}
      \begin{subfigure}[b]{0.3\textwidth}
         \centering
         \includegraphics[width=\textwidth]{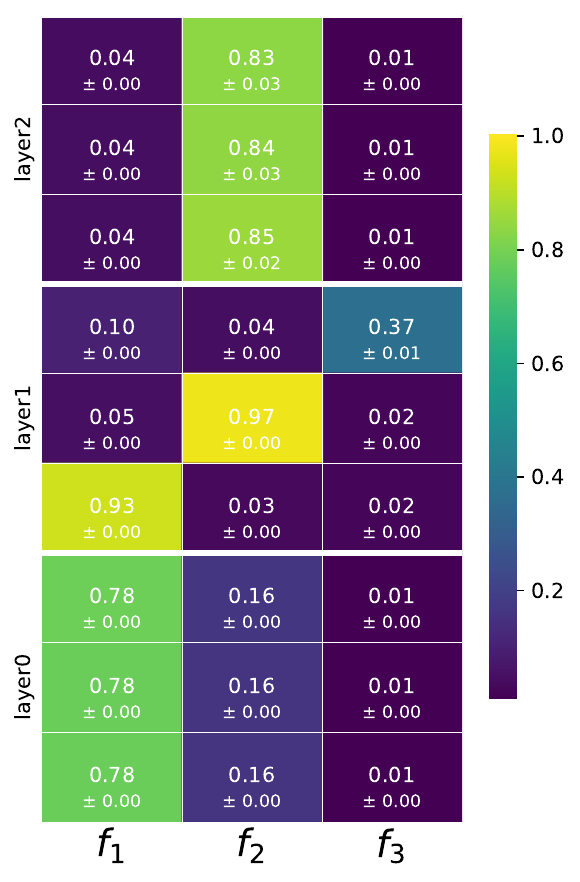}
         \caption{CKA scores}
     \end{subfigure}
    \caption{Forecast and CKA scores of the attention bottleneck Autoformer on synthetic data, where the three heads of each layer (vertically) are compared with the three concept vectors (horizontally).} 
    \label{fig:synthetic main}
\end{figure}

Each concept in the bottleneck is defined as one of the underlying functions (i.e., $f_1$, $f_2$ or $f_3$), for which the ground-truth is known by construction. For hyperparameter $\alpha=0.8$ (see Section \ref{sec:loss}), we find that the model is able to forecast well, while achieving very high similarity scores. That is, the model obtains a Mean Squared Error (MSE) of 0.36 ± 0.17 and Mean Absolute Error (MAE) of 0.46 ± 0.12 on 5 different seeds. See Figure \ref{fig:synthetic main} for a sample forecast on the test data and the CKA scores of the model's representations with the concept representations. The heads in the bottleneck \texttt{layer1} show high similarity for their respective concepts, e.g. a score of 0.93 for the head trained on $f_1$ (recall that CKA scores range from 0 for totally dissimilar to 1.0 for identical, although potentially scaled and rotated). We refer to Appendix \ref{app:synthetic} for more results on the synthetic dataset.

\subsection{Real-world data}
Table \ref{tab:performance analysis} shows the performance of the Autoformer with our bottleneck on the benchmark datasets, compared to the AR surrogate model (i.e. the first interpretable concept) and \citet{wu_autoformer_2021} (i.e. the original Autoformer model). Note that the bottleneck models are trained with a free component, i.e., $c=3$, and the original Autoformer is of a different size (two encoder layers with eight heads per layer). Visualizations of the forecasts from these models are shown in Appendix \ref{app: qualitative results}.

\floatsetup[table]{objectset=centering,capposition=top}
    \begin{table}[h!]
        \caption{Error scores of different Autoformer models. For both metrics, it holds that a lower score indicates a better performance, where the best results are \textbf{bold}, and the second-best are \underline{underlined}. }\label{tab:performance analysis}
        \resizebox{\textwidth}{!}{
        \begin{tabular}{lrrrrrrrr|rr}\toprule
            &\multicolumn{2}{c}{\textbf{Att bottleneck}}&\multicolumn{2}{c}{\textbf{FF bottleneck}}&\multicolumn{2}{c}{\textbf{No bottleneck}}&\multicolumn{2}{c|}{\textbf{AR}}&\multicolumn{2}{c}{\textbf{Wu et al.}}
            \\\cmidrule(r){2-3}\cmidrule(r){4-5}\cmidrule(r){6-7}\cmidrule(r){8-9}\cmidrule(r){10-11}  
            &MSE&MAE&MSE&MAE&MSE&MAE&MSE&MAE&MSE&MAE\\\midrule

Electricity    & 0.231                   & ~0.338 & \uline{0.207}          & \uline{0.320}  & 0.280                  & 0.368         & 0.497          & 0.522          & \textbf{0.201~} & \textbf{0.317~} \\
Traffic        & 0.642                   & 0.393  & \textbf{0.393}         & \textbf{0.377} & 0.619                  & \uline{0.387} & \uline{0.420}  & 0.494          & 0.613           & 0.388           \\
Weather        & 0.290                   & 0.354  & 0.271                  & 0.341          & 0.269                  & 0.344         & \textbf{0.006} & \textbf{0.062} & \uline{0.266}   & \uline{0.336}   \\
Illness        & 3.586                   & 1.313  & 3.661                  & 1.322          & \uline{3.405}          & 1.295         & \textbf{1.027} & \textbf{0.820} & 3.483           & \uline{1.287}   \\
Exchange rate  & 0.195                   & 0.323  & 0.155                  & 0.290          & \uline{0.152}          & 0.283         & \textbf{0.082} & \textbf{0.230} & 0.197           & \uline{0.323}   \\
ETT            & 0.177                   & 0.282  & 0.174                  & 0.280          & \uline{0.155}          & \uline{0.265} & \textbf{0.034} & \textbf{0.117} & 0.255           & 0.339           
            \\\bottomrule
        \end{tabular}}
    \end{table} 

We find that including a bottleneck (either \textbf{Att bottleneck} or \textbf{FF bottleneck}) outperforms \textbf{Wu et al.} for three datasets (traffic, exchange rate and ETT), and stays within 5\% of the MSE and MAE for the other three datasets.
Surprisingly, the surrogate  AR model outperforms the other models for most datasets w.r.t. both MSE and MAE, even though this model is very simple.\footnote{Note that the phenomenon that simple models sometimes beat time series transformers \citep{zeng_are_2022} has been observed before. There has been a vivid discussion about the relevance of these results, for instance \href{https://huggingface.co/blog/autoformer}{here}. These discussions are beyond the scope of our paper, which rather targets interpretability of time series transformers. For more information on the effect of AR as surrogate model, see Appendix \ref{app:ar}.}  
More detailed results are presented in Appendix \ref{app: detailed results} and \ref{app: sensitivity}, where the first includes the results for bottlenecks without free component (including the standard deviation for different seeds), and the latter includes a sensitivity analysis to hyperparameter $\alpha$.

Similar to the Autoformer, the Vanilla Transformer and FEDformer with a bottleneck outperform models without bottleneck for some datasets, see Appendix \ref{app: framework extension} and \ref{app: framework extension FEDformer} for a full analysis, respectively. 

\subsection{Interpretability Analysis}
To demonstrate the impact of the bottleneck on model interpretability, we first conduct a CKA analysis on the bottleneck layer with the corresponding interpretable concepts, and then visually demonstrate how each component contributes to the final forecast.

\subsubsection{CKA Analysis}
To test the extent to which the bottleneck represents the interpretable concepts, we calculate the CKA scores of the model's representations with the concept representations. The scores of the feed-forward bottleneck on the electricity dataset are shown in Figure \ref{fig:CKA scores} (see Appendix \ref{app: sensitivity} for more scores on the Autoformer). Note that the bottom, middle and upper layer of \texttt{layer1} correspond to the AR, hour-of-day, and free component of the bottleneck, respectively. 

The scores show that the representations in the bottleneck layer are much more similar to the intended concepts than the representations from the model without bottleneck: 0.94 for the AR model, and 1.00 for the hour-of-day feature, whereas the model without bottleneck does not show high similarity to the interpretable concepts. This indicates that the training framework can encourage the components to form representations that are perfectly similar to the interpretable concepts. Additionally, note that the CKA scores of other layers than the bottleneck layer are also higher in Fig. \ref{fig: ff bottleneck cka}, which indicates that these other model components also learn to represent the interpretable concepts. This does not affect the interpretability of the bottleneck layer itself (Section \ref{ch: intervention}).

\begin{figure}[h!]
     \centering
     \begin{subfigure}[b]{0.35\textwidth}
         \centering
         \includegraphics[width=\textwidth]{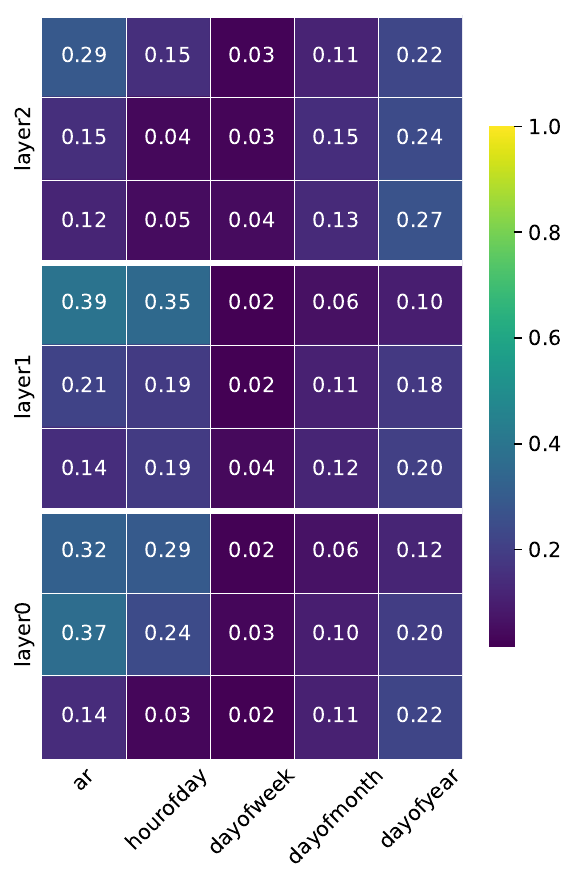}
         \caption{No bottleneck}
         \label{fig:alpha is 0}
     \end{subfigure}
      \begin{subfigure}[b]{0.35\textwidth}
         \centering
         \includegraphics[width=\textwidth]{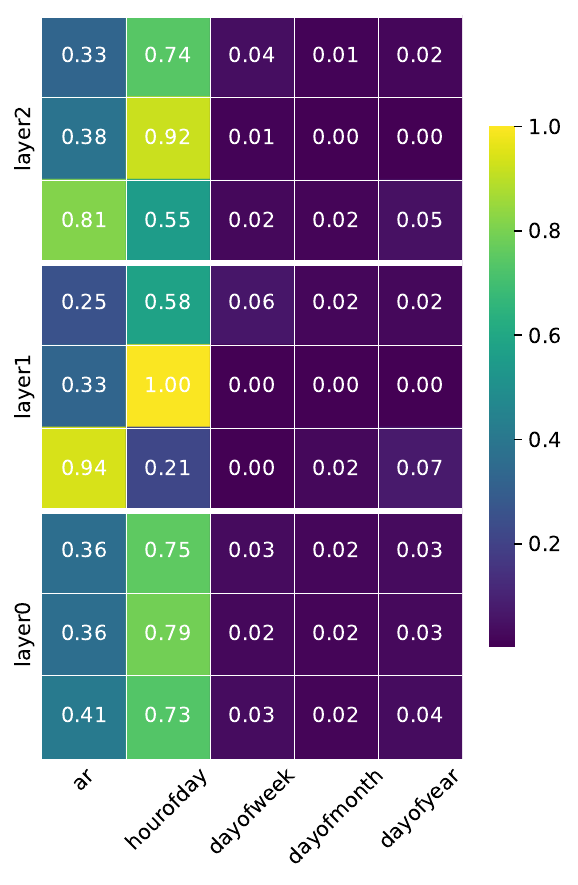}
         \caption{Ours (FF bottleneck)}
         \label{fig: ff bottleneck cka}
     \end{subfigure}
     
    \caption{CKA scores on different concepts for the encoder of the Vanilla Transformer without bottleneck and with FF bottleneck. Both models contain three heads per layer. The first component of \texttt{layer1} (lower row) of the attention bottleneck is trained to be similar to AR, and the second component (middle row) to the hour-of-day concept. The scores are calculated on three batches of size 32 from the electricity test data. Recall that CKA is defined on a scale from 0 to 1, where 1 denotes perfect similarity.}
    \label{fig:CKA scores}
\end{figure}

\subsubsection{Component Visualizations}
\label{sec:compvisualizations}

\begin{figure}[h!]
     \centering
     \begin{subfigure}[b]{0.32\textwidth}
         \centering
         \includegraphics[width=\textwidth]{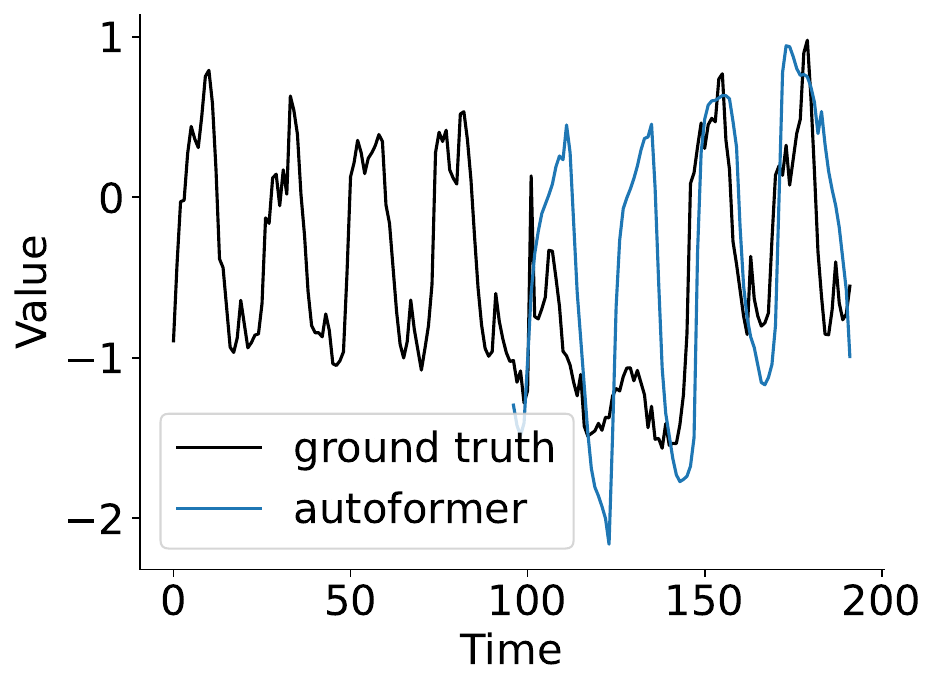}
         \caption{Comp. 1 (AR)}
         \label{fig: comp 0}
     \end{subfigure}
     \hfill
      \begin{subfigure}[b]{0.32\textwidth}
         \centering
         \includegraphics[width=\textwidth]{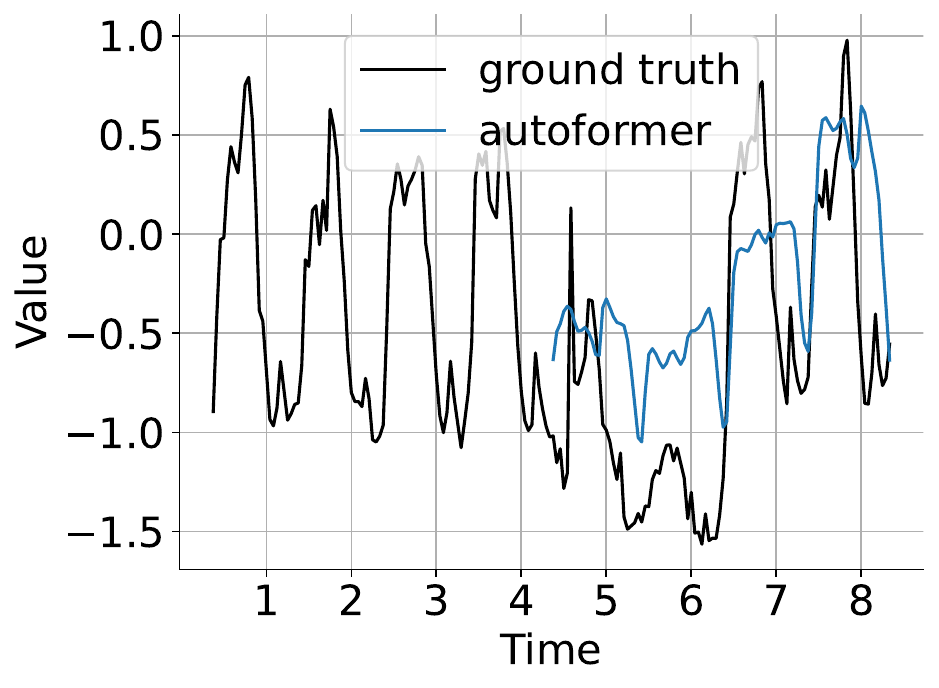}
         \caption{Comp. 2 (Hour-of-day)}
         \label{fig: comp 1}
     \end{subfigure}
     \hfill
    \begin{subfigure}[b]{0.32\textwidth}
        \centering
        \includegraphics[width=\textwidth]{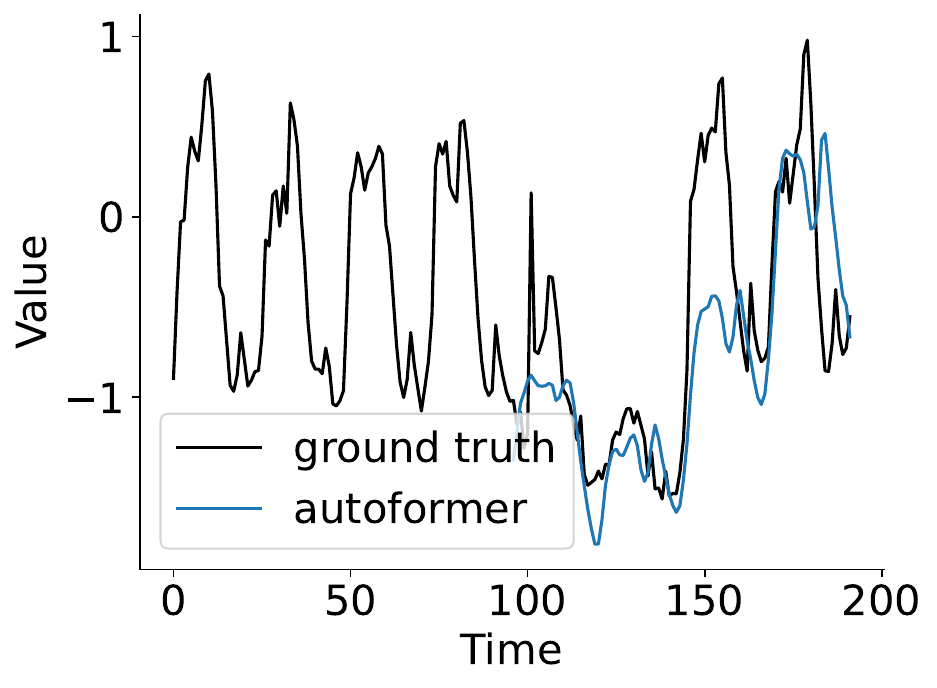}
        \caption{Comp. 3 (Side-channel)}
        \label{fig: comp 2}
    \end{subfigure}
    \begin{subfigure}[b]{0.32\textwidth}
        \centering
        \includegraphics[width=\textwidth]{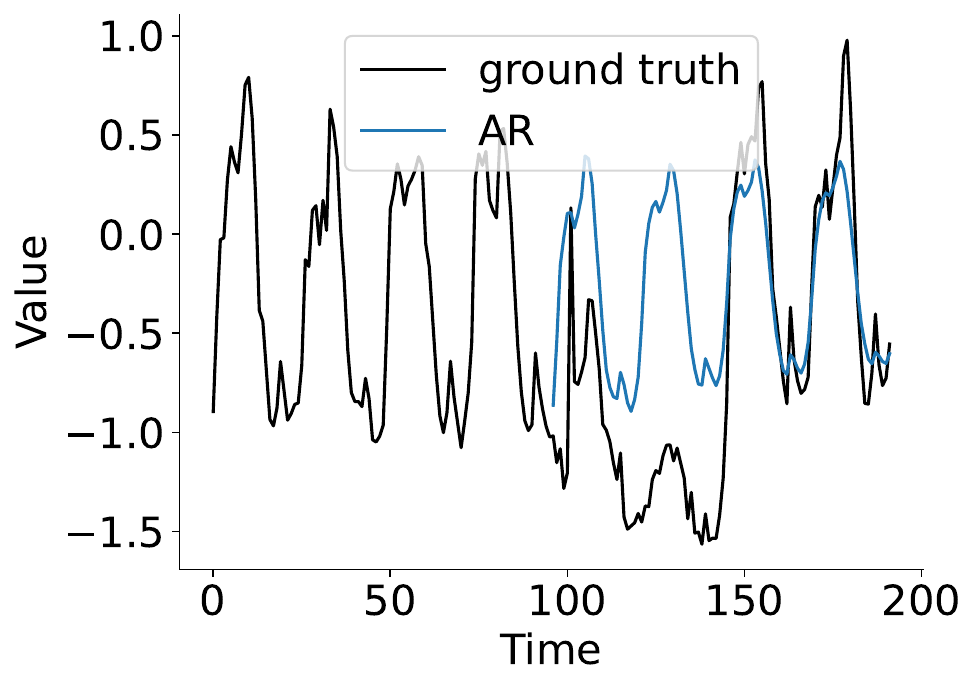}
        \caption{AR}
        \label{fig: ar}
    \end{subfigure}
    \hfill 
    \begin{subfigure}[b]{0.32\textwidth}
        \centering
        \includegraphics[width=\textwidth]{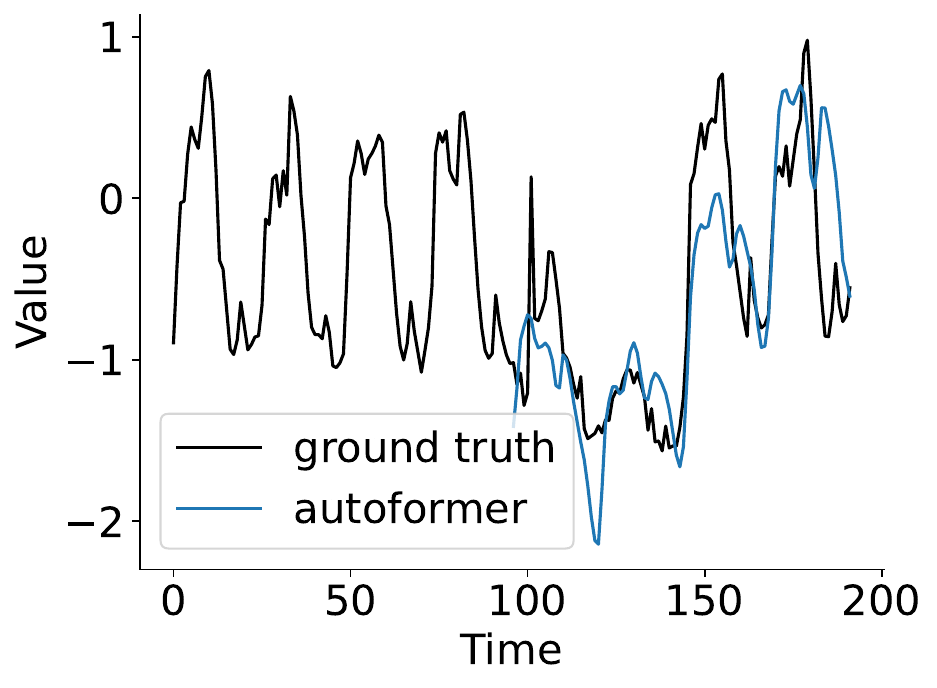}
        \caption{Full bottleneck}
        \label{fig: full bottleneck}
    \end{subfigure}
    \hfill 
    \begin{subfigure}[b]{0.32\textwidth}
        \centering
        \includegraphics[width=\textwidth]{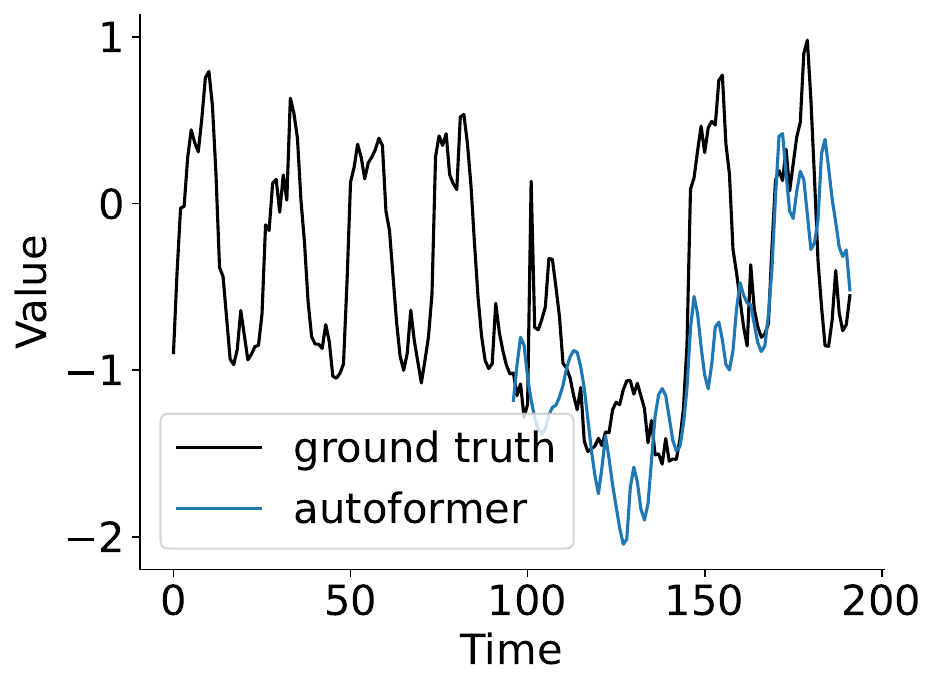}
        \caption{Final, using only comp. 3}
        \label{fig: extra comp3 final}
    \end{subfigure}
     
    \caption{Forecasts from individual bottleneck components by masking the other components with zero in \ref{fig: comp 0}, \ref{fig: comp 1} and \ref{fig: comp 2} (FF bottleneck Autoformer on electricity data). The first half of the ground truth forms the input to the model. Note that the horizontal axes are the same across all figures, but Figure \ref{fig: comp 1} contains a grid of days instead of numbered hours. Figure \ref{fig: ar} shows the forecast made by the surrogate model AR; Figure \ref{fig: full bottleneck} shows the forecast of the entire layer (i.e., all components together), and \ref{fig: extra comp3 final} shows the forecast of the final layer when only the third component is used in the bottleneck layer. Note the difference between Figures \ref{fig: comp 2} and \ref{fig: extra comp3 final}, where we decode from the bottleneck and the final layer, respectively.}
    \label{fig:comp visualizations}
\end{figure}

Because the model components all read and write from the residual stream \citep{elhage_mathematical_2021}, we can visualize the contributions to the final prediction of each component separately by applying the entire decoder to the component representations (Decoder Lens method, \citet{langedijk_decoderlens_2023}). This way, we obtain visualizations of the contributions of each component in the bottleneck, see Figure \ref{fig:comp visualizations}. We obtain the output from the full bottleneck by applying the decoder to the output of the bottleneck (after performing layer normalization). The output from each component individually is obtained by masking the other components with zero (close to the mean). 

From Figure \ref{fig: comp 0} and \ref{fig: comp 1} we see that the different bottleneck components are similar to the concepts they were trained on. In particular, the first component shows a forecast with correct periodicity and few irregularities, similar to the actual forecast from the AR model. Likewise, the second component shows a periodicity to the actual hour-of-day feature. The third component is not trained to be similar to an interpretable concept, and seems to pick up on the high-frequency patterns in the data, e.g., the low, second peak in the forecast. This observation is further strengthened by Figure \ref{fig: extra comp3 final}, which shows that the final forecast consists of many high-frequency patterns when using only the third component from the bottleneck. We find similar component visualizations on the Vanilla Transformer, see Appendix \ref{app: vanilla component}.

\subsection{Intervention} \label{ch: intervention}

The main benefit of interpreting trained models is gaining a deeper understanding and, possibly, more control of the model's behavior. This can be useful in the scenario of out-of-distribution data at inference time. If the data changes in features that can be interpreted in the model, it is feasible to intervene locally in these concepts to exclusively employ the model with data from its training distribution. Additionally, an intervention can be regarded as a causal interpretability test, where a successful intervention indicates a successful representation of the concept of interest.

To show such benefit of our framework, we perform activation patching (or causal tracing, \citealp{meng_locating_2023}), where causal effects of hidden state activations are researched by evaluating the model on clean and corrupted inputs. We evaluate the trained model on data with shifted timestamps and compare it with performing an intervention on the shifted concept.

\begin{figure}[]
\captionsetup{}
    \centering
    \includegraphics[width=.9\linewidth]{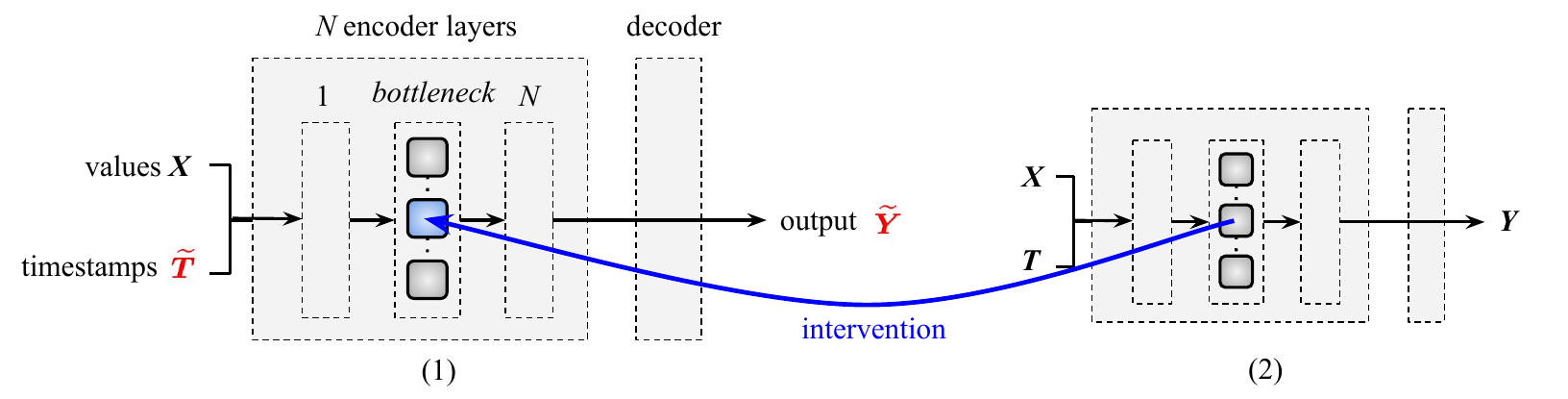}
    \caption{Intervention experiment, where we run the model on time-shifted input (1), but replace the activations of the hour-of-day component with those from a run (2) on unshifted timestamps $\mT$.
    }
    \label{fig:overview intervention}
\end{figure}

More specifically, we delay the input timestamps $\mT \in \mathbb{R}^{I \times 4}$ with a fixed number of hours to obtain the shifted timestamps $\widetilde{\mT}$, so that the learned patterns associated to the hour-of-day feature are misleading. We run the model on both types of timestamps, and perform an intervention in the bottleneck by substituting the activations based on the shifted time with the activations based on the original, see Figure \ref{fig:overview intervention} for an overview.

\color{black}

\begin{figure}[h!]
    \centering


    \begin{subfigure}[b]{0.4\textwidth}
        \centering
        \includegraphics[width=\textwidth]{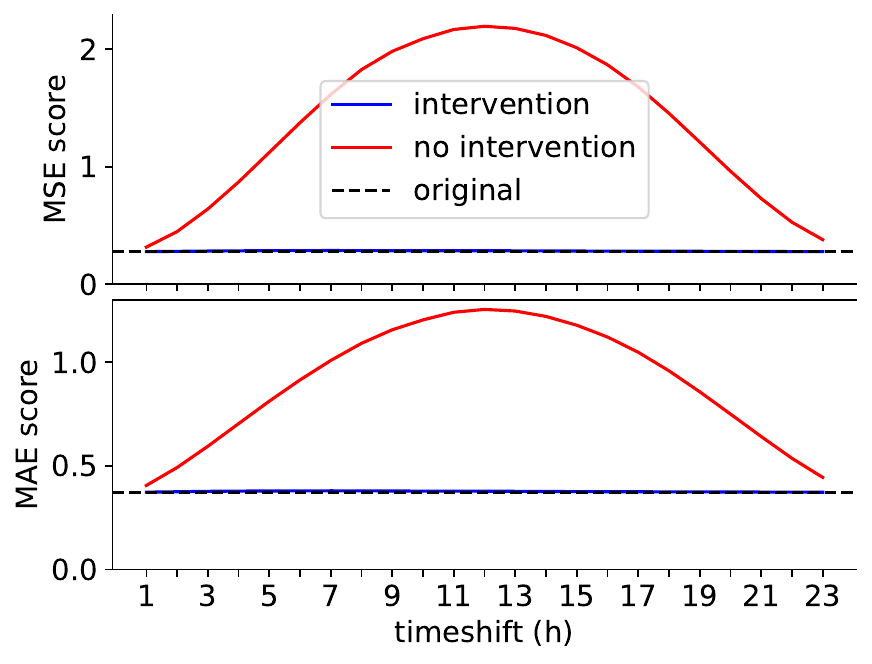}
        \caption{V. Transformer - Att}
        \label{fig:att bottleneck}
    \end{subfigure}
    \begin{subfigure}[b]{0.4\textwidth}
        \centering
        \includegraphics[width=\textwidth]{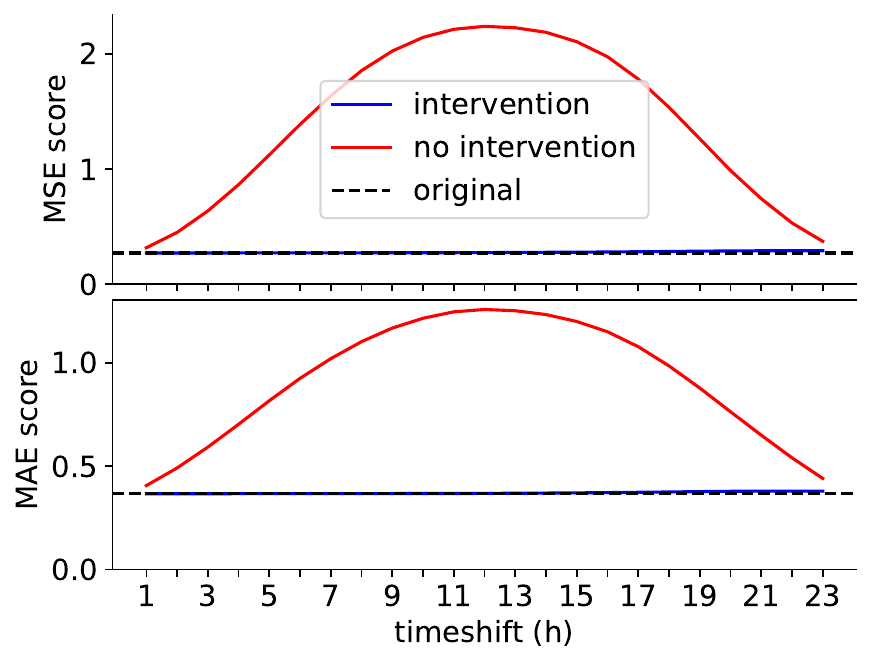}
        \caption{V. Transformer - FF}
        \label{fig:ff bottleneck}
    \end{subfigure}
    \vspace{1em}  
    \par
    \begin{subfigure}[b]{0.4\textwidth}
        \centering
        \includegraphics[width=\textwidth]{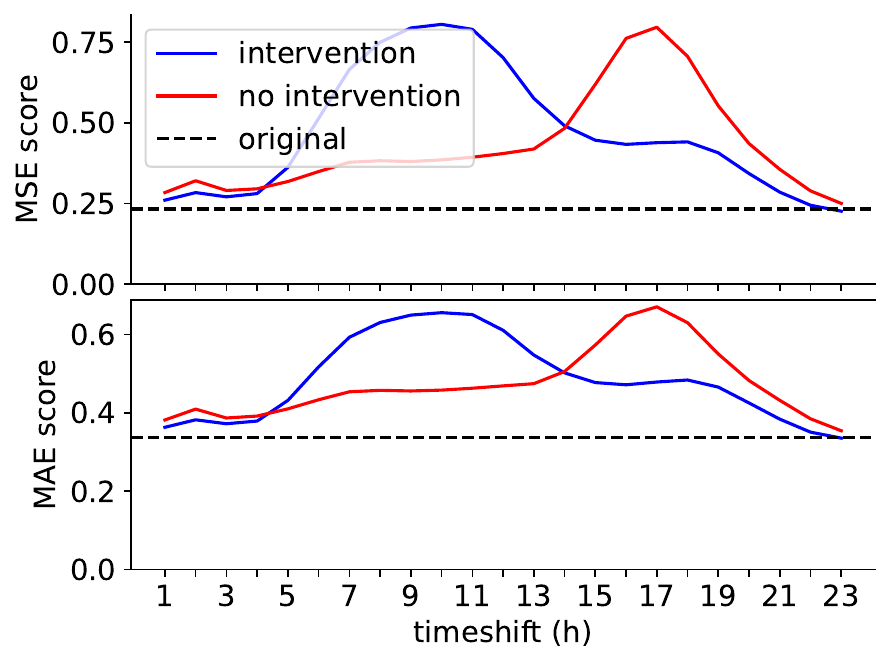}
        \caption{Autoformer - Att}
        \label{fig:att bottleneck auto}
    \end{subfigure}
    \begin{subfigure}[b]{0.4\textwidth}
        \centering
        \includegraphics[width=\textwidth]{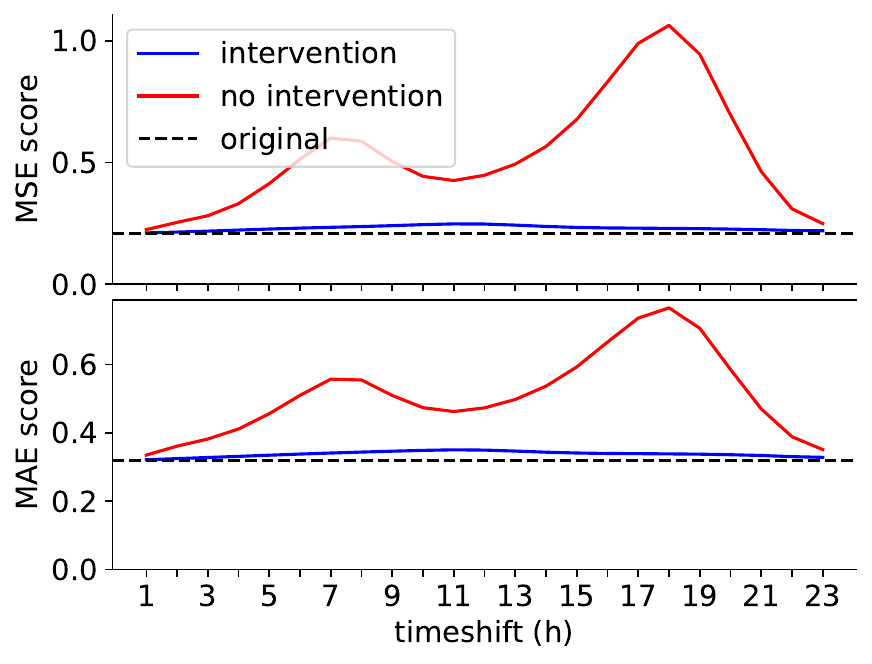}
        \caption{Autoformer - FF}
        \label{fig:ff bottleneck auto}
    \end{subfigure}

     
    \caption{MSE of the attention (Att) and feed-forward (FF) bottleneck models on electricity data with shifted timestamps. The dashed line represents the performance on the data without timeshift.}
    \label{fig:intervention}
\end{figure}

We perform the intervention experiment with the electricity dataset, and perform shifts of up to and including 23 hours. We compare the performance of the intervention with out-of-the-box performance of the same model on the shifted dataset. The results of the Vanilla Transformer shown in Figure \ref{fig:intervention}. 
 
Remarkably, the intervention on the Vanilla Transformer achieves the original performance for all timeshifts. 
This indicates that the bottleneck models effectively learn to represent the hour-of-day concept in the dedicated bottleneck component. Most interestingly, the models only utilise this interpretable concept in the bottleneck layer, but not in other encoder layers (because the experiment only intervenes in the bottleneck).



\section{Discussion and Conclusions} \label{sect: discussion}

In this work, we propose a training framework based on Concept Bottleneck Models to enforce interpretability of time series transformers. We introduce a new loss function based on the similarity score CKA of the model's representations and interpretable concepts. We apply our framework to the Vanilla Transformer, Autoformer and FEDformer using synthetic data and six benchmark datasets. Our results indicate that the overall performance remains unaffected, while the model's components become more interpretable. Additionally, it becomes possible to perform a local intervention when employing the model after a temporal data shift. 

The main limitation of our concept bottleneck framework is that interpretable concepts have to be decided on before training, which might require domain knowledge. Representations for these concepts have to be available during training. However, domain-agnostic concepts such as the AR surrogate model and hour-of-day information are sufficient. Additionally, our framework increases computational complexity. This might be problematic if the size of the architecture increases.

An interesting direction for future research would be to optimize the number and type of interpretable concepts in the bottleneck, and extend the framework to other modalities. We trained mostly using two domain-agnostic concepts (AR and hour-of-day), but including more concepts, possibly domain-specific, would be very interesting. For example, one could consider choosing speech and music concepts for audio time series. Additionally, the framework should also work for transformers in other modalities, e.g., language and vision, although these models are usually of larger size. We hope our work contributes to a deeper understanding of (time series) transformers and their behavior in different fields. In particular, recent progress in the field of mechanistic interpretability is based on the observation that the residual stream of the transformer encourages modular solutions, which enables localized concepts or specialized circuitry to perform a specific task. Instead of relying on post-hoc localization of these concepts, our paper presents a demonstration that we can encourage locality of concepts, without a significant loss in performance.

Regarding societal impact, this work enables transparent time series forecasting models, which enable explainable forecasts. However, in the case of malicious use, biases could be included in the models. Harm could be prevented by developing mechanistic interpretability techniques for bias detection in time series models. 
\newpage
\bibliography{references}
\bibliographystyle{plainnat}


\appendix

\section{Datasets} \label{app: datasets}
We evaluate the Autoformer model on six real-world benchmarks, covering the five domains of energy, traffic, economics, weather, and disease. We use the same datasets as \citet{wu_autoformer_2021}, and provide additional information in Table \ref{tab: datasets}, as given in the original Autoformer paper.

\begin{table}[h!]
\caption{Descriptions of the datasets, as given by \citet{wu_autoformer_2021} and shared \href{https://drive.google.com/drive/folders/1ZOYpTUa82_jCcxIdTmyr0LXQfvaM9vIy?usp=sharing}{online}. `Pred len' denotes the prediction length used in our experiments.} \label{tab: datasets}
\begin{tabularx}{\textwidth}{llX}
\hline
Dataset & Pred len & Description \\ 
\hline
Electricity & 96 & Hourly electricity consumption of 321 customers from 2012 to 2014. \ \\
Traffic & 96 & Hourly data from California Department of Transportation, which describes the road occupancy rates measured by different sensors on San Francisco Bay area freeways. \\
Weather & 96 & Recorded every 10 minutes for 2020 whole year, which contains 21 meteorological indicators, such as air temperature, humidity, etc. \\
Illness & 24 & Includes the weekly recorded influenza-like illness (ILI) patients data from Centers for Disease Control and Prevention of the United States between 2002 and 2021, which describes the ratio of patients seen with ILI and the total number of the patients. \\
Exchange rate & 96 & Daily exchange rates of eight different countries ranging from 1990 to 2016. \\
ETT & 96 & Data collected from electricity transformers, including load and oil temperature that are recorded every~15 minutes between July 2016 and July 2018. \\
\hline
\end{tabularx}
\end{table}

\section{Formalization of Concept Bottleneck Framework} \label{app: formalization}
Any time series Transformer obtains two types of input: (1) \emph{data values} $\mX \in \mathbb{R}^{I \times d}$, and (2) \emph{timestamps} $\mT \in \mathbb{R}^{I \times 4}$.
The transformer consists of an encoder and a decoder, which are both constructed from one or multiple layers. Any encoder layer contains two sub-layers: a multi-head attention mechanism (Att) and a fully connected neural network (FF). Every sub-layer contains a residual connection around it. More specifically, the output $\mX^{\ell}$ of any encoder layer $\ell$ is:
\begin{align*}
    \mX^{\ell} &= \textrm{Encoder}(\mX^{\ell-1}) \\
    &= \textrm{LayerNorm}(\textrm{FF}(\mS^{\ell}) + \mS^{\ell}), \\
    \mS^{\ell} &= \textrm{LayerNorm}(\textrm{Att}(\mX^{\ell-1}) + \mX^{\ell-1}), 
\end{align*}

where 
\begin{align*}
    \textrm{FF}(\rvx) &= \max(0, \, \rvx \mW_1 +\rvb_1) \, \mW_2 + \rvb_2, \\
    \textrm{Att}(\rvx) &= \mW_0 \cdot \textrm{Concat}\left(\textrm{h}_1(\rvx), \, \dots, \, \textrm{h}_h(\rvx) \right).
\end{align*}


For future reference, we denote the output of the feed-forward module as follows: $\textrm{FF}(\mS^\ell) = \mZ^\ell\in \mathbb{R}^{d_1 \times d_2}$. 
We omit the definition of the decoder, because our bottleneck framework does not include it. Note that the exact implementation of each (sub-)layer depends on the type of Transformer. 

\subsection{Bottleneck Layer}
We assign one encoder layer to be the bottleneck and construct it such that it contains $c$ latent representations or \textit{components}, i.e., $\left(\mH_{i}\right)_{i=1}^c$. Depending on the bottleneck type $\tau$, these latent representations are either taken from the attention mechanism or the feed-forward module. More specifically:
\begin{equation*}
    \mH_{i} = \begin{cases} 
\textrm{h}_i(\rvx) & \text{if bottleneck type } \tau = \text{Att}, \\
\mZ_{i} & \text{if bottleneck type } \tau = \textrm{FF}. \\
\end{cases}
\end{equation*}

Since the attention block is multi-headed, different heads naturally form the components of the attention bottleneck. For the feed-forward bottleneck, we define the components to be slices (in $d_1$) from its output $\mZ$, such that stacking the components results in the original output. 

Note that the residual connection around the corresponding bottleneck component is removed, and that each component $\mH_{i}$ should represent a pre-defined interpretable concept. 


\subsection{Intervention}

In the intervention experiment, we shift the time stamps $\mT$ to obtain $\widetilde{\mT}$. The key aspect of the experiment is to run the Transformer on the shifted time stamps $\widetilde{\mT}$, and replace the input representations $\widetilde{\mX}^{b-1}$ of the bottleneck layer $b$ with ${\mX}^{b-1}$ (based on $\mT$), but only in the component that represents the time concept. 

More specifically, if type $\tau = \textrm{Att}$, we intervene on the attention block in the bottleneck as follows:
\begin{equation*}
    \textrm{Att}(\rvx, \widetilde{\rvx}) = \mW_0 \cdot \textrm{Concat}\left(\textrm{h}_1(\widetilde{\rvx}), \, \textrm{h}_2(\rvx), \, \textrm{h}_3(\widetilde{\rvx}) \right),
\end{equation*}

and, if type $\tau = \textrm{FF}$, as follows:
\begin{equation*}
    \textrm{FF}(\rvx, \widetilde{\rvx}) = \textrm{Stack}(\widetilde{\mZ}_1, \mZ_2, \widetilde{\mZ}_3).
\end{equation*}

In both functions we make use of the fact that the time concept is represented in the second component, and there are three components in total. This intervention can be done in the bottleneck only, because, by construction, its location of the concept representations is known.

\newpage
\section{Qualitative Results} \label{app: qualitative results}

\begin{figure}[h!]
     \centering
     \begin{subfigure}[b]{0.47\textwidth}
         \centering
         \includegraphics[width=\textwidth]{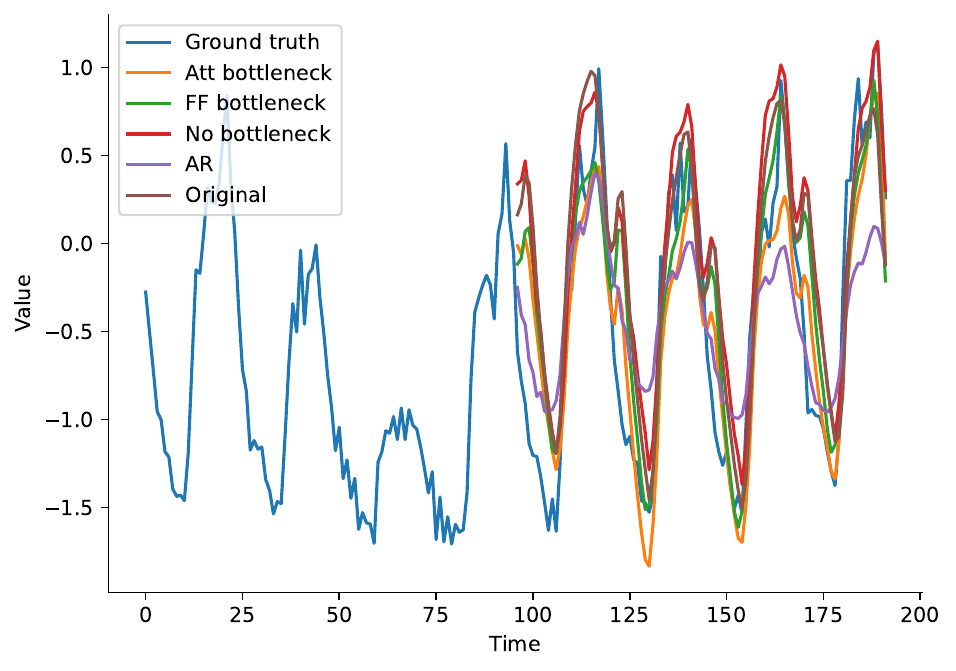}
         \caption{Electricity}
         \label{}
     \end{subfigure}
     \begin{subfigure}[b]{0.47\textwidth}
         \centering
         \includegraphics[width=\textwidth]{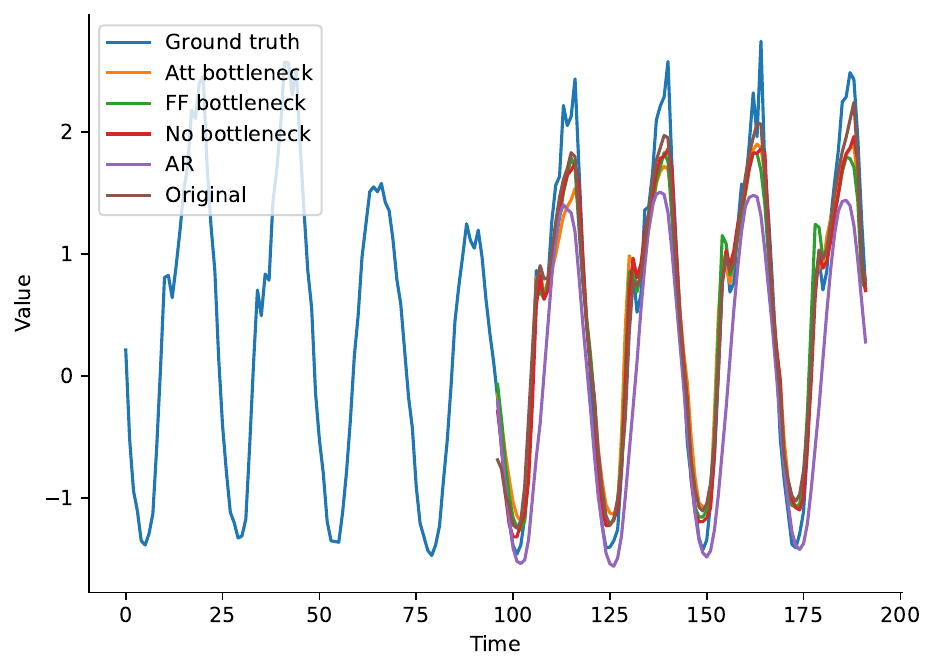}
         \caption{Traffic}
         \label{}
     \end{subfigure}
     \begin{subfigure}[b]{0.47\textwidth}
         \centering
         \includegraphics[width=\textwidth]{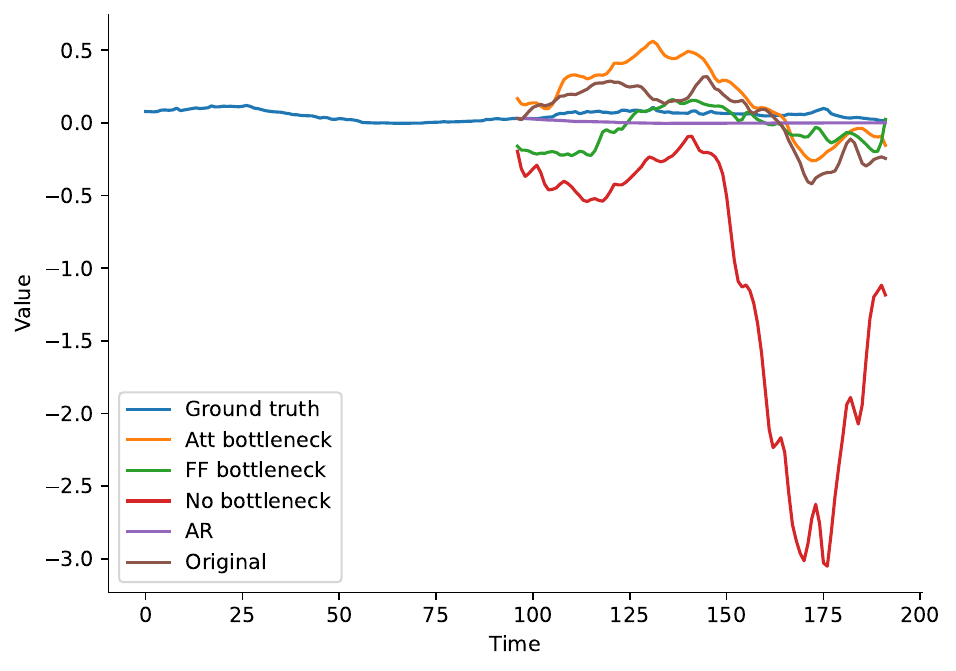}
         \caption{Weather}
         \label{}
     \end{subfigure}
     \begin{subfigure}[b]{0.47\textwidth}
         \centering
         \includegraphics[width=\textwidth]{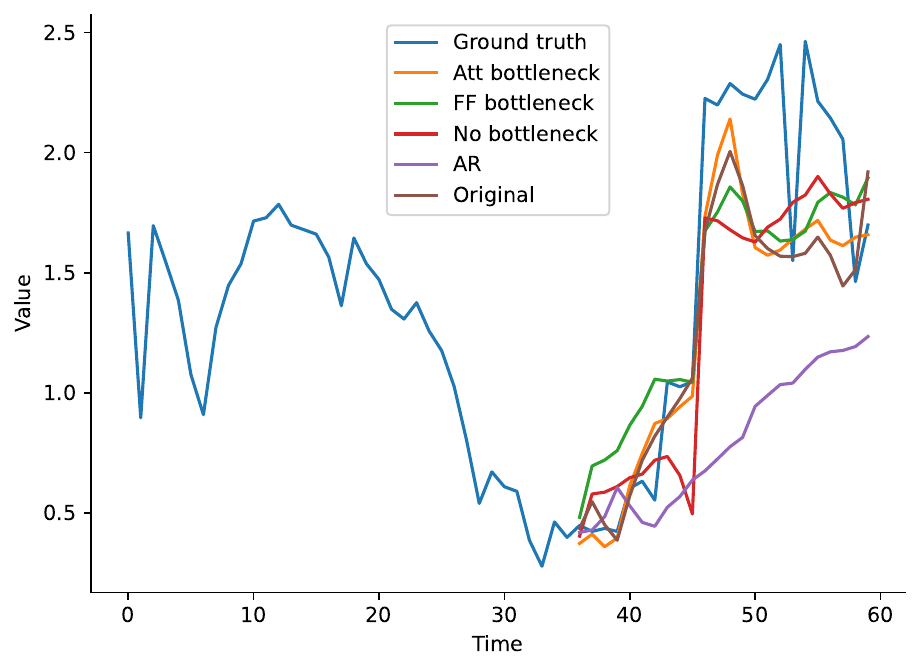}
         \caption{Illness}
         \label{}
     \end{subfigure}
     \begin{subfigure}[b]{0.47\textwidth}
         \centering
         \includegraphics[width=\textwidth]{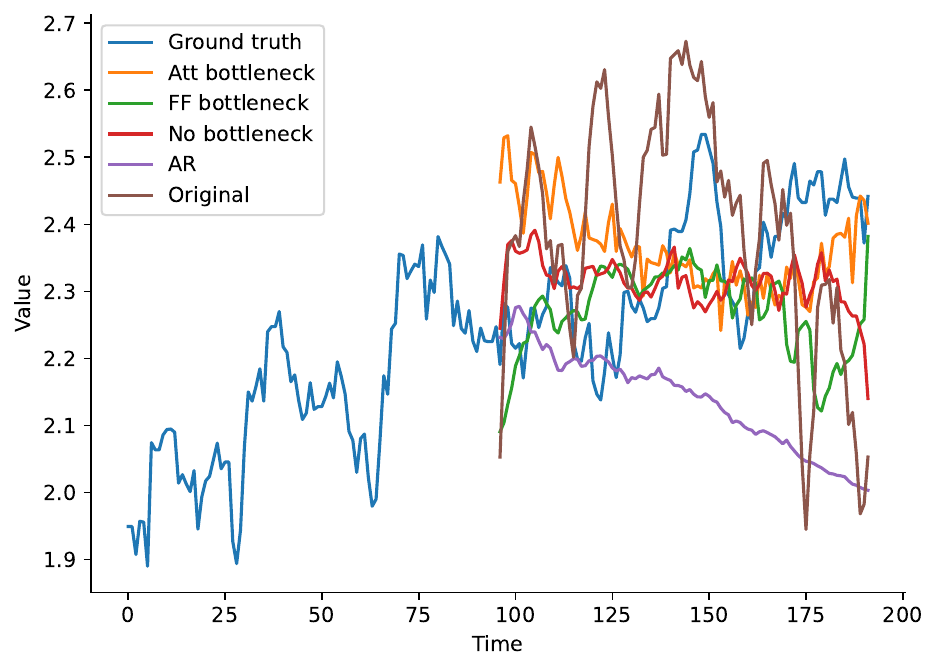}
         \caption{Exchange rate}
         \label{}
     \end{subfigure}
     \begin{subfigure}[b]{0.47\textwidth}
         \centering
         \includegraphics[width=\textwidth]{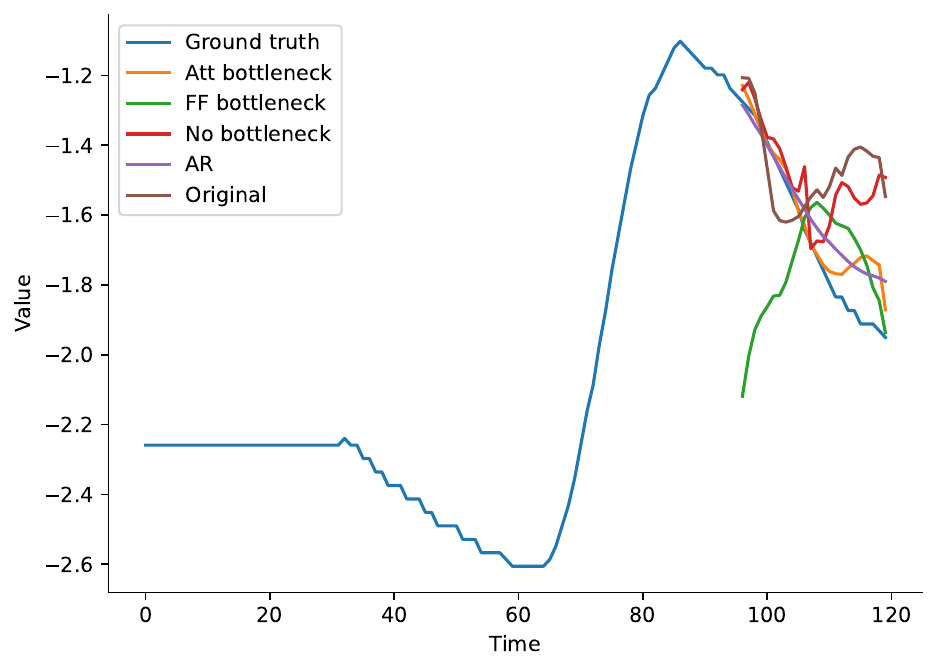}
         \caption{ETT}
         \label{}
     \end{subfigure}
        \caption{Forecasts on different datasets. The first part of the ground truth (shown in blue) is the input for the models, and the test set is used for each dataset.}
        \label{fig:app-qualitative results}
\end{figure}

\newpage
\section{Detailed results} \label{app: detailed results}

\floatsetup[table]{objectset=centering,capposition=top}
\begin{table}[h!]
\centering
\caption{Performance of different models in Mean Squared Error (MSE) and Mean Absolute Error (MAE). The bottlenecks \textbf{do} contain a free component ($c=3$), and use AR as surrogate model. The model with no bottleneck is an original Autoformer of similar size. For all datasets, the shortest prediction lengths from \cite{wu_autoformer_2021} are used, see Table \ref{tab: datasets}. The standard deviation is determined using five different seeds.}
\label{tab:app performance free head}
\resizebox{\textwidth}{!}{
\begin{tblr}{
  row{2} = {c},
  cell{1}{1} = {r=2}{},
  cell{1}{2} = {c=2}{c},
  cell{1}{4} = {c=2}{c},
  cell{1}{6} = {c=2}{c},
  cell{1}{8} = {c=2}{c},
  cell{1}{10} = {c=2}{c},
  cell{3}{2} = {c},
  cell{3}{3} = {c},
  cell{3}{4} = {c},
  cell{3}{5} = {c},
  cell{3}{6} = {c},
  cell{3}{7} = {c},
  cell{3}{8} = {c},
  cell{3}{9} = {c},
  cell{3}{10} = {c},
  cell{3}{11} = {c},
  cell{4}{2} = {c},
  cell{4}{3} = {c},
  cell{4}{4} = {c},
  cell{4}{5} = {c},
  cell{4}{6} = {c},
  cell{4}{7} = {c},
  cell{4}{8} = {c},
  cell{4}{9} = {c},
  cell{4}{10} = {c},
  cell{4}{11} = {c},
  cell{5}{2} = {c},
  cell{5}{3} = {c},
  cell{5}{4} = {c},
  cell{5}{5} = {c},
  cell{5}{6} = {c},
  cell{5}{7} = {c},
  cell{5}{8} = {c},
  cell{5}{9} = {c},
  cell{5}{10} = {c},
  cell{5}{11} = {c},
  cell{6}{2} = {c},
  cell{6}{3} = {c},
  cell{6}{4} = {c},
  cell{6}{5} = {c},
  cell{6}{6} = {c},
  cell{6}{7} = {c},
  cell{6}{8} = {c},
  cell{6}{9} = {c},
  cell{6}{10} = {c},
  cell{6}{11} = {c},
  cell{7}{2} = {c},
  cell{7}{3} = {c},
  cell{7}{4} = {c},
  cell{7}{5} = {c},
  cell{7}{6} = {c},
  cell{7}{7} = {c},
  cell{7}{8} = {c},
  cell{7}{9} = {c},
  cell{7}{10} = {c},
  cell{7}{11} = {c},
  cell{8}{2} = {c},
  cell{8}{3} = {c},
  cell{8}{4} = {c},
  cell{8}{5} = {c},
  cell{8}{6} = {c},
  cell{8}{7} = {c},
  cell{8}{8} = {c},
  cell{8}{9} = {c},
  cell{8}{10} = {c},
  cell{8}{11} = {c},
  hline{1,9} = {-}{0.08em},
  hline{2} = {2-11}{l},
  hline{3} = {1}{},
  hline{3} = {2-11}{0.03em},
  vline{10} = {3-8}{},
}
Free component & \textbf{Att bottleneck} &                & \textbf{FF bottleneck} &               & \textbf{No bottleneck} &                & \textbf{AR}    &                &  \textbf{Wu et al.} &       \\
               & MSE                     & MAE            & MSE                    & MAE           & MSE                    & MAE            & MSE            & MAE            & MSE               & MAE   \\
Electricity    & {0.231 \\± 0.009~}      & {~0.338 \\± 0.005} & {\uline{0.207 }\\± 0.005}  & {\uline{0.320}\\± 0.005}   & {0.280 \\± 0.165}         & {0.368 \\± 0.111}         & 0.497          & 0.522          & {\textbf{0.201~}\\± 0.003} & {\textbf{0.317~}\\± 0.004} \\
Traffic        & {0.642 \\± 0.022}       & {0.393 \\± 0.013}  & {\textbf{0.393 }\\± 0.013} & {\textbf{0.377 }\\± 0.006} & {0.619 \\± 0.015}         & {\uline{0.387 }\\± 0.005} & \uline{0.420}  & 0.494          & {0.613 \\± 0.028}          & {0.388 \\± 0.012}          \\
Weather        & {0.290 \\± 0.027}       & {0.354 \\± 0.020}  & {0.271 \\± 0.016}          & {0.341 \\± 0.011}          & {0.269 \\± 0.000}         & {0.344 \\± 0.000}         & \textbf{0.006} & \textbf{0.062} & {\uline{0.266 }\\± 0.007}  & {\uline{0.336 }\\± 0.006}  \\
Illness        & {3.586 \\± 0.241}       & {1.313 \\± 0.040}  & {3.661 \\± 0.237}          & {1.322 \\± 0.050}          & {\uline{3.405 }\\± 0.208} & {1.295 \\± 0.044}         & \textbf{1.027} & \textbf{0.820} & {3.483\\± 0.107}           & {\uline{1.287}\\± 0.018}       \\
Exchange rate  & {0.195 \\± 0.029}       & {0.323 \\± 0.025}  & {0.155 \\± 0.010}          & {0.290 \\± 0.013}          & {\uline{0.152 }\\± 0.003} & {0.283 \\± 0.003}         & \textbf{0.082} & \textbf{0.230} & {0.197 \\± 0.019}          & {\uline{0.323 }\\± 0.012}  \\
ETT            & {0.177 \\± 0.003}       & {0.282 \\± 0.004}  & {0.174 \\± 0.006}          & {0.280 \\± 0.005}          & {\uline{0.155 }\\± 0.004} & {\uline{0.265 }\\± 0.002} & \textbf{0.034} & \textbf{0.117} & {0.255 \\± 0.020}          & {0.339 \\± 0.020}          
\end{tblr}}
\end{table}

\floatsetup[table]{objectset=centering,capposition=top}
\begin{table}[htbp!]
\centering
\caption{Performance on different datasets, where the bottlenecks \textbf{do not} contain a free component \textbf{($c=2$)}. AR is used as surrogate model in the bottlenecks. The model with no bottleneck is an original Autoformer of similar size. For all datasets, the shortest prediction lengths from \cite{wu_autoformer_2021} are used, see Table \ref{tab: datasets}. The standard deviation is determined using five different seeds.}
\label{tab:app performance no free head}
\resizebox{\textwidth}{!}{
\begin{tblr}{
  row{2} = {c},
  cell{1}{1} = {r=2}{},
  cell{1}{2} = {c=2}{c},
  cell{1}{4} = {c=2}{c},
  cell{1}{6} = {c=2}{c},
  cell{1}{8} = {c=2}{c},
  cell{1}{10} = {c=2}{c},
  cell{3}{2} = {c},
  cell{3}{3} = {c},
  cell{3}{4} = {c},
  cell{3}{5} = {c},
  cell{3}{6} = {c},
  cell{3}{7} = {c},
  cell{3}{8} = {c},
  cell{3}{9} = {c},
  cell{3}{10} = {c},
  cell{3}{11} = {c},
  cell{4}{2} = {c},
  cell{4}{3} = {c},
  cell{4}{4} = {c},
  cell{4}{5} = {c},
  cell{4}{6} = {c},
  cell{4}{7} = {c},
  cell{4}{8} = {c},
  cell{4}{9} = {c},
  cell{4}{10} = {c},
  cell{4}{11} = {c},
  cell{5}{2} = {c},
  cell{5}{3} = {c},
  cell{5}{4} = {c},
  cell{5}{5} = {c},
  cell{5}{6} = {c},
  cell{5}{7} = {c},
  cell{5}{8} = {c},
  cell{5}{9} = {c},
  cell{5}{10} = {c},
  cell{5}{11} = {c},
  cell{6}{2} = {c},
  cell{6}{3} = {c},
  cell{6}{4} = {c},
  cell{6}{5} = {c},
  cell{6}{6} = {c},
  cell{6}{7} = {c},
  cell{6}{8} = {c},
  cell{6}{9} = {c},
  cell{6}{10} = {c},
  cell{6}{11} = {c},
  cell{7}{2} = {c},
  cell{7}{3} = {c},
  cell{7}{4} = {c},
  cell{7}{5} = {c},
  cell{7}{6} = {c},
  cell{7}{7} = {c},
  cell{7}{8} = {c},
  cell{7}{9} = {c},
  cell{7}{10} = {c},
  cell{7}{11} = {c},
  cell{8}{2} = {c},
  cell{8}{3} = {c},
  cell{8}{4} = {c},
  cell{8}{5} = {c},
  cell{8}{6} = {c},
  cell{8}{7} = {c},
  cell{8}{8} = {c},
  cell{8}{9} = {c},
  cell{8}{10} = {c},
  cell{8}{11} = {c},
  hline{1,9} = {-}{0.08em},
  hline{2} = {2-11}{l},
  hline{3} = {1}{},
  hline{3} = {2-11}{0.03em},
  vline{10} = {3-8}{},
}
No free component & \textbf{Att bottleneck} &                & \textbf{FF bottleneck} &               & \textbf{No bottleneck} &                & \textbf{AR}    &                &  \textbf{Wu et al.} &       \\
               & MSE                     & MAE            & MSE                    & MAE           & MSE                    & MAE            & MSE            & MAE            & MSE               & MAE   \\
Electricity       & {0.224\\± 0.006}       & {0.332 \\± 0.003} & {0.206 \\± 0.009}      & {0.321 \\± 0.009} & {\uline{0.202 }\\± 0.006} & {\uline{0.318 }\\± 0.007}  & 0.497          & 0.522          & {\textbf{0.201}\\± 0.003} & {\textbf{0.317}\\± 0.004} \\
Traffic           & {0.629 \\± 0.023}       & {0.394 \\± 0.015} & {0.627 \\± 0.031}      & {0.392 \\± 0.025} & {\uline{0.613 }\\± 0.018} & {\textbf{0.378 }\\± 0.007} & \textbf{0.420} & 0.494          & {0.613 \\± 0.028}          & {\uline{0.388}\\± 0.012}  \\
Weather           & {0.281 \\± 0.025}       & {0.348 \\± 0.018} & {0.260\\± 0.015}       & {0.333 \\± 0.013} & {\uline{0.257 }\\± 0.004} & {\uline{0.332 }\\± 0.005}  & \textbf{0.006} & \textbf{0.062} & {0.266 \\± 0.007}          & {0.336 \\± 0.006}          \\
Illness           & {3.966 \\± 0.296}       & {1.401 \\± 0.073} & {3.721 \\± 0.268}      & {1.351 \\± 0.053} & {3.585 \\± 0.331}         & {1.333 \\± 0.070}           & \textbf{1.027} & \textbf{0.820} & {\uline{3.483} \\ ± 0.107}      & {\uline{1.287}\\± 0.018}      \\
Exchange rate     & {0.208 \\± 0.026}       & {0.333 \\± 0.022} & {0.158 \\± 0.009}      & {0.293 \\± 0.009} & {\uline{0.152 }\\± 0.006} & {\uline{0.284 }\\± 0.007}  & \textbf{0.082} & \textbf{0.230} & {0.197 \\± 0.019}          & {0.323 \\± 0.012}          \\
ETT               & {0.178 \\± 0.011}       & {0.283 \\± 0.007} & {0.174 \\± 0.01}       & {0.283 \\± 0.009} & {\uline{0.165 }\\± 0.004} & {\uline{0.274 }\\± 0.004}  & \textbf{0.034} & \textbf{0.117} & {0.255 \\± 0.020}          & {0.339 \\± 0.020}          
\end{tblr}}
\end{table}

\newpage
\section{Hyper-Parameter Sensitivity} \label{app: sensitivity}
\begin{figure}[htbp!]
     \centering
     \includegraphics[width=0.5\textwidth]{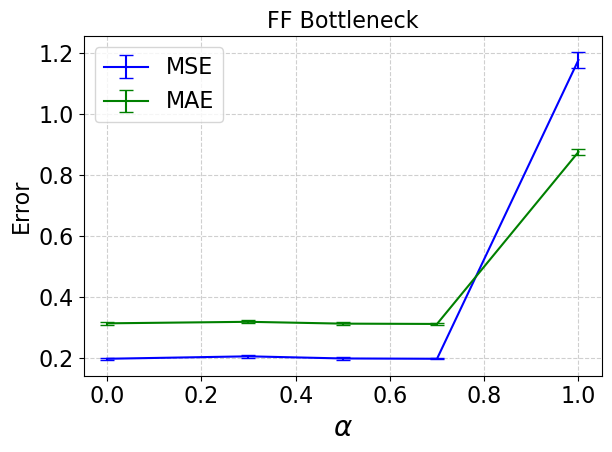}
     \caption{Performance of the Autoformer for different values of $\alpha$ in MSE and MAE.}
     \label{fig:app alphas}
\end{figure}

To verify the sensitivity to hyperparameter $\alpha$ in the loss function, we train the Autoformer with a feed-forward bottleneck on different values for $\alpha$, where the bottleneck contains a free component ($c=3$) and the model is trained on the electricity dataset. The results are given in Figure \ref{fig:app alphas}. Interestingly, the error scores for all $\alpha < 1$ are close in value, which verifies that additionally training for interpretability does not hurt the performance, at least not in this set-up. Note that a low forecasting error cannot be expected for $\alpha=1$, because in this edge case the loss function does not contain any term that represents the forecasting performance.




\begin{figure}[htbp!]
     \centering
     \begin{subfigure}[b]{0.4\textwidth}
         \centering
         \includegraphics[width=\textwidth]{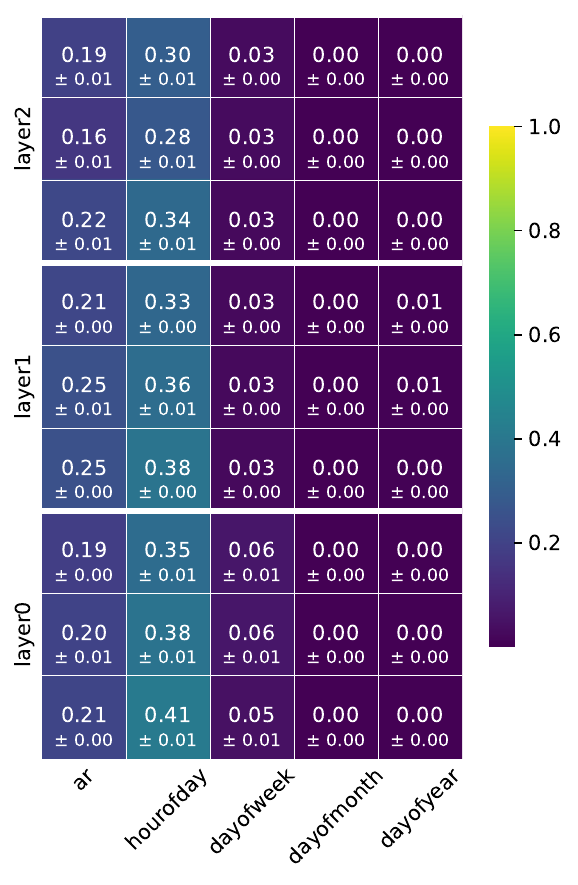}
         \caption{$\alpha=0$}
         \label{fig:alpha is 0}
     \end{subfigure}
     \begin{subfigure}[b]{0.4\textwidth}
         \centering
         \includegraphics[width=\textwidth]{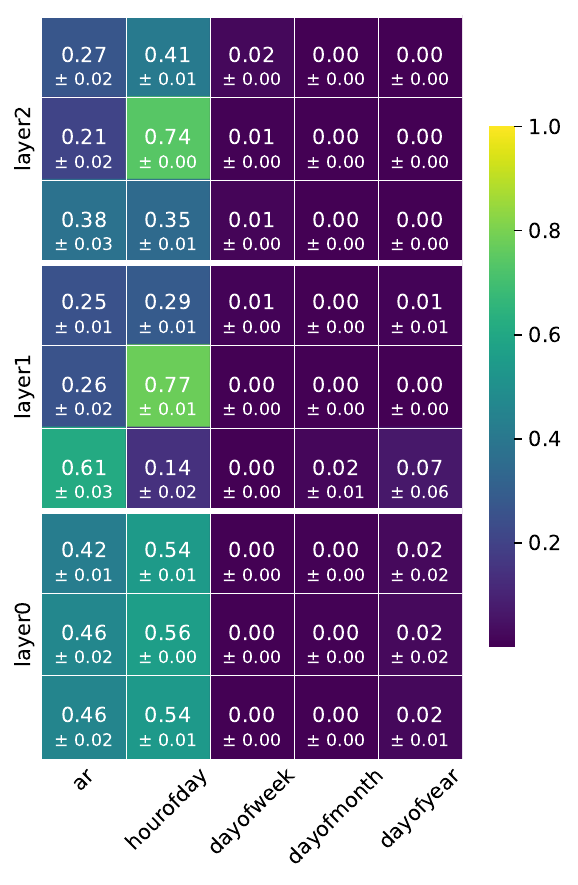}
         \caption{$\alpha=0.3$}
         \label{}
     \end{subfigure}
    \caption{CKA scores of the feed-forward bottleneck Autoformer on electricity data for different values of hyperparameter $\alpha$. The scores are calculated using three batches of size 32 of the test data set.}
    \label{fig:app alpha 0}
\end{figure}

\begin{figure}[htbp!]
     \centering
     \begin{subfigure}[b]{0.4\textwidth}
         \centering
         \includegraphics[width=\textwidth]{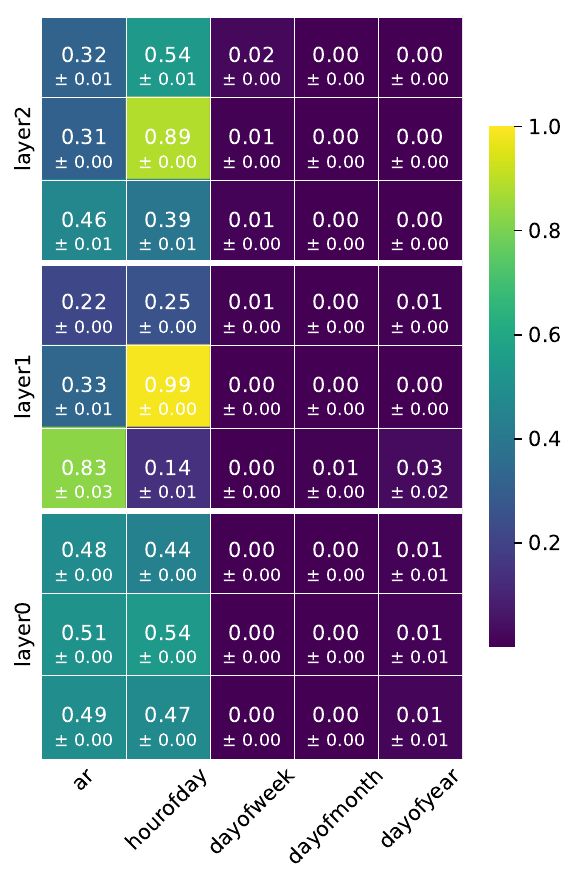}
         \caption{$\alpha=0.5$}
         \label{}
     \end{subfigure}
     \begin{subfigure}[b]{0.4\textwidth}
         \centering
         \includegraphics[width=\textwidth]{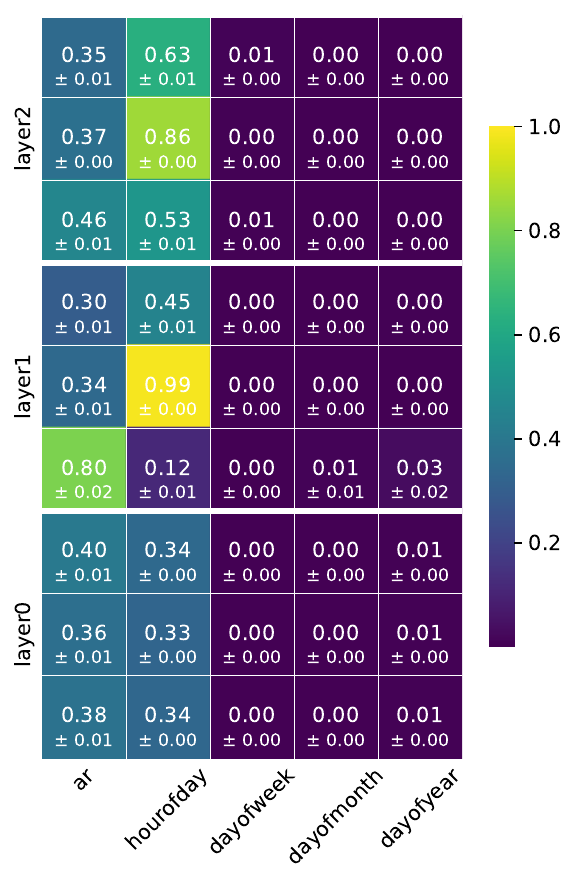}
         \caption{$\alpha=0.7$}
         \label{}
     \end{subfigure}
    \caption{CKA scores of the feed-forward bottleneck Autoformer on electricity data for different values of hyperparameter $\alpha$. The scores are calculated using three batches of size 32 of the test data set.}
    \label{fig:app alpha 1}
\end{figure}

\begin{figure}[htbp!]
     \centering
     \begin{subfigure}[b]{0.4\textwidth}
         \centering
         \includegraphics[width=\textwidth]{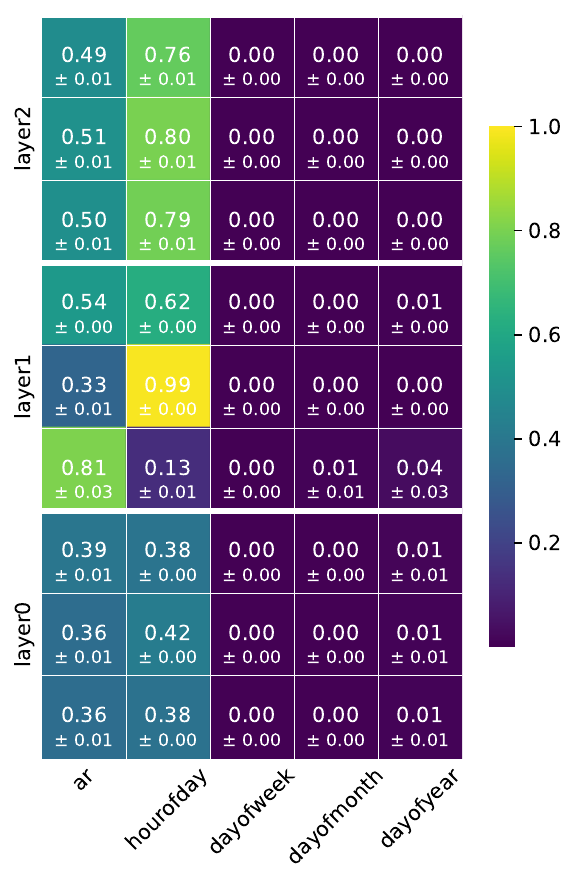}
         \caption{$\alpha=1$}
         \label{}
     \end{subfigure}
    \caption{CKA scores of the feed-forward bottleneck Autoformer on electricity data for hyperparameter $\alpha=1$. The scores are calculated using three batches of size 32 of the test data set.}
    \label{fig:app alpha 2}
\end{figure}

Additionally, the CKA scores of the different models with the interpretable concepts (and other time features) are given in Figures \ref{fig:app alpha 0}, \ref{fig:app alpha 1}, and \ref{fig:app alpha 2}. Naturally, the CKA scores are the lowest in the setting $\alpha=0$, and the scores from the bottleneck (\texttt{layer1}) increase over $\alpha$. Interestingly, the CKA scores from the bottleneck do not increase for higher values than $\alpha=0.5$, although the scores of some other components do increase. This indicates that perfect similarity (i.e. CKA score of 1) to some interpretable concepts may not be reached.

     
\newpage
\section{Application of Framework to Vanilla Transformer} \label{app: framework extension}
To demonstrate the generality of the concept bottleneck framework, we apply it to an additional Transformer architecture, namely the \emph{vanilla Transformer} (the original architecture from which all Transformer models, including all time series Transformers, are derived). We train it using the same six benchmark datasets and perform a similar, but less extensive, analysis as done for the Autoformer model. Note that the architecture of the Transformer is \textit{not} modified, and the timestamps are included as an embedding (in addition to the positional embedding).

\subsection{Performance Analysis} \label{app: vanilla performance}
The performance of the vanilla Transformer model with and without bottleneck is given in Table \ref{tab:performance analysis transformer}. We train the bottleneck with a `free' component (the side channel), i.e., with $c=3$. Note that \cite{wu_autoformer_2021} do not provide scores for these benchmark forecasting datasets, therefore we cannot include them in the table. The results show that the vanilla Transformer performs, unsurprisingly, worse than the Autoformer, and for most datasets also worse than the linear AR model. However, most relevant, for our purposes, is that across the datasets using a concept bottleneck does not hurt the overall performance of the vanilla Transformer.

\floatsetup[table]{objectset=centering,capposition=top}
    \begin{table}[h!]
    \caption{Performance of different vanilla Transformer models. For both metrics, it holds that a lower score indicates a better performance, where the best results are \textbf{bold}, and the second-best are \underline{underlined}. }\label{tab:performance analysis transformer}
        \begin{tabular}{lrrrrrrrr}\toprule
            &\multicolumn{2}{c}{\textbf{Att bottleneck}}&\multicolumn{2}{c}{\textbf{FF bottleneck}}&\multicolumn{2}{c}{\textbf{No bottleneck}}&\multicolumn{2}{c}{\textbf{AR}}
            \\\cmidrule(r){2-3}\cmidrule(r){4-5}\cmidrule(r){6-7}\cmidrule(r){8-9}  
            &MSE&MAE&MSE&MAE&MSE&MAE&MSE&MAE\\\midrule
Electricity   & \underline{0.275}       & \underline{0.371}       & \textbf{0.268}       & \textbf{0.362}      & \underline{0.275}                & \underline{0.371}                & 0.497          & 0.522          \\
Traffic       & 0.708                & \underline{0.394}       & 0.703                & 0.397               & \underline{0.684}       & \textbf{0.376}       & \textbf{0.420} & 0.494          \\
Weather       & 0.400                & 0.450                & 0.381                & \underline{0.410}      & \underline{0.362}       & 0.415                & \textbf{0.006} & \textbf{0.062} \\
Illness       & 3.380                & 1.280                & 3.323                & \underline{1.252}      & \underline{3.321}       & 1.273                & \textbf{1.027} & \textbf{0.820} \\
Exchange rate & \underline{0.675}       & 0.642                & 0.677                & \underline{0.633}      & 0.694                & 0.662                & \textbf{0.082} & \textbf{0.230} \\
ETT           & 0.230                & 0.328                & 0.185                & 0.299               & \underline{0.166}       & \underline{0.294}       & \textbf{0.034} & \textbf{0.117}            
            \\\bottomrule
        \end{tabular}
    \end{table}

\subsection{CKA Analysis} \label{app: vanilla CKA}
\begin{figure}[htbp!]
     \centering
     
     \begin{subfigure}[b]{0.35\textwidth}
         \centering \includegraphics[width=\textwidth]{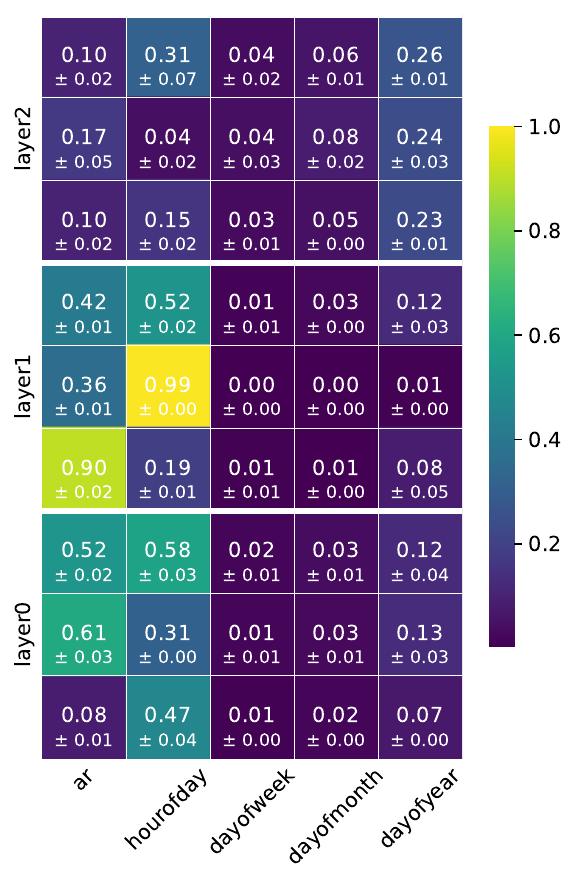}
         \caption{Att bottleneck}
         \label{}
     \end{subfigure}
     \begin{subfigure}[b]{0.35\textwidth}
         \centering
         \includegraphics[width=\textwidth]{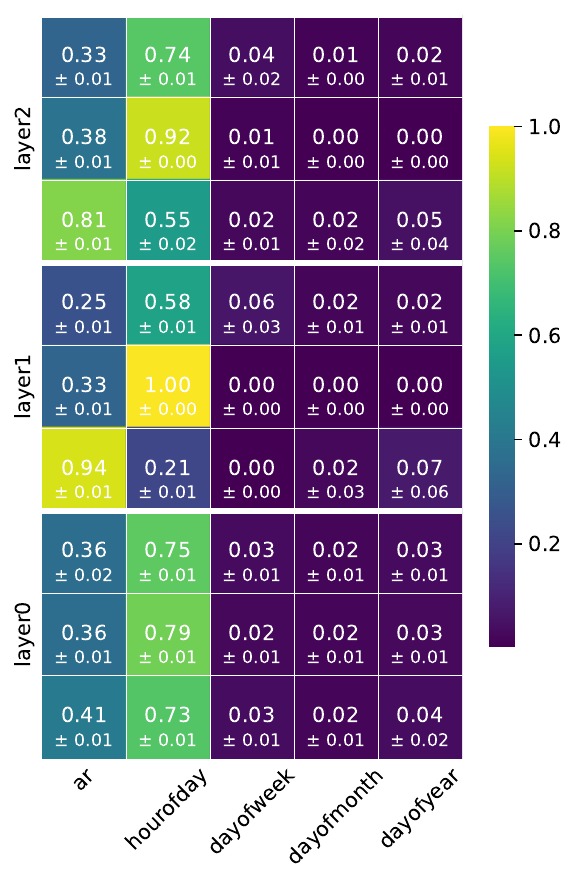}
         \caption{FF bottleneck}
         \label{}
     \end{subfigure}
    \caption{CKA scores of the vanilla Transformer's encoder (containing three heads per layer) from the attention and feed-forward bottleneck on the electricity dataset, where each score denotes the similarity of an individual component. The first component of \texttt{layer1}  is trained to be similar to AR, and the second component to the hour-of-day concept (lower and middle row in the figure, respectively). The scores are calculated using three batches of size 32 from the test data set.}
    \label{fig:CKA scores transformer}
\end{figure}

After training the vanilla Transformer with the bottleneck framework, we evaluate the similarity of its hidden representations to the interpretable concepts using CKA, see Figure \ref{fig:CKA scores transformer}. Recall that CKA scores are defined in the range from 0 to 1, where 1 indicates perfect similarity. Both components in the two types of bottleneck show very high similarity to their target concept. Interestingly, the first component in the bottleneck (the AR concept) shows a higher similarity to the AR representations than the Autoformer (see Figure \ref{fig:CKA scores}), presumably because the decomposition structure of the Autoformer hinders learning a linear function.

\subsection{Component Visualizations} \label{app: vanilla component}
\begin{figure}[h!]
     \centering
     
     \begin{subfigure}[b]{0.3\textwidth}
         \centering
         \includegraphics[width=\textwidth]{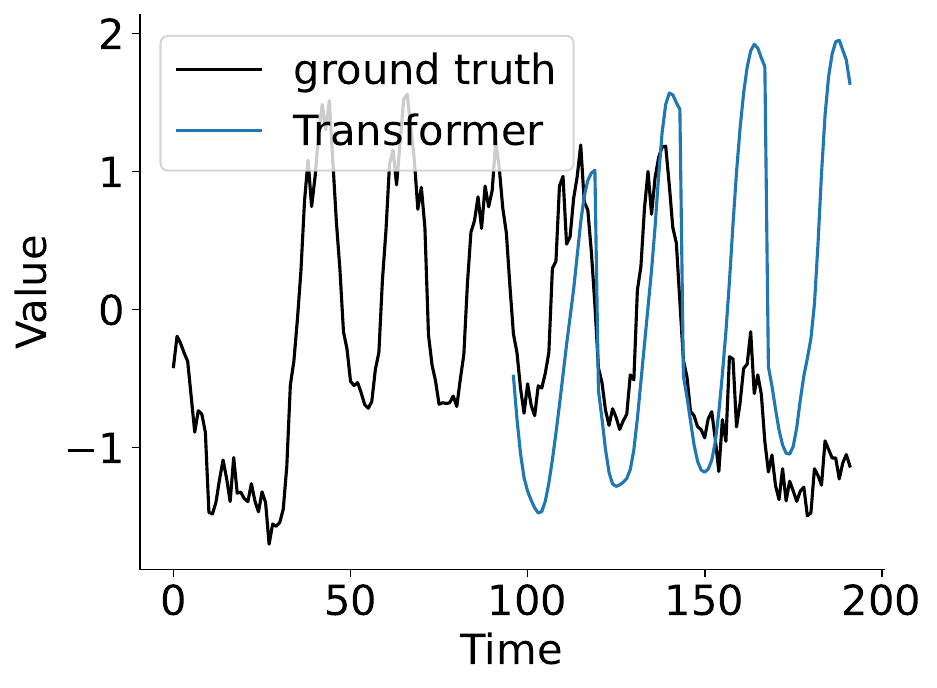}
         \caption{Comp. 1 (`AR')}
         \label{fig: comp 0 trans}
     \end{subfigure}
     \begin{subfigure}[b]{0.3\textwidth}
         \centering
         \includegraphics[width=\textwidth]{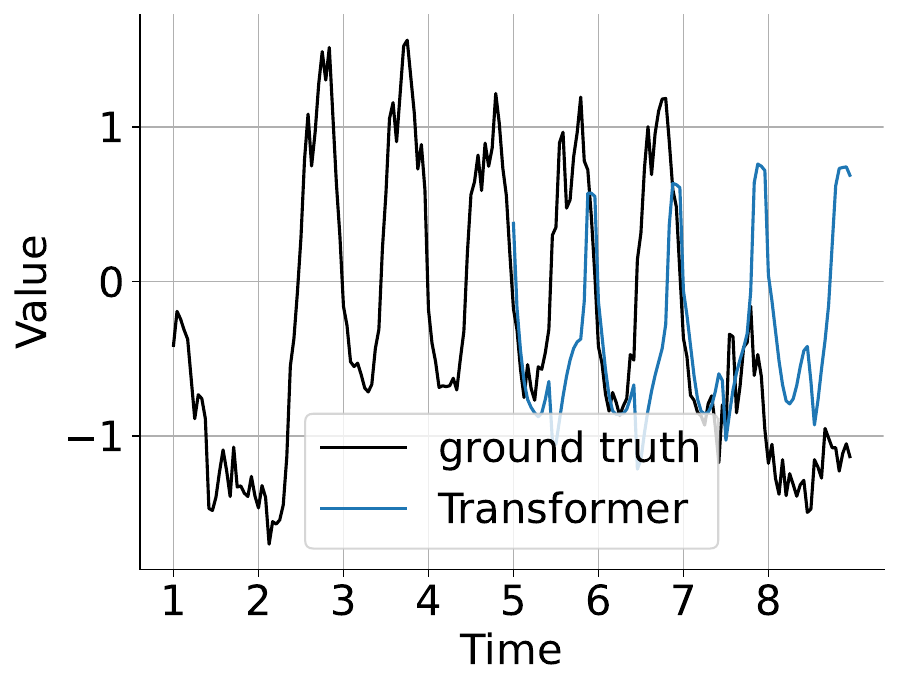}
         \caption{Comp. 2 (`Hour-of-day')}
         \label{fig: comp 1 trans}
     \end{subfigure}
     \begin{subfigure}[b]{0.3\textwidth}
         \centering
         \includegraphics[width=\textwidth]{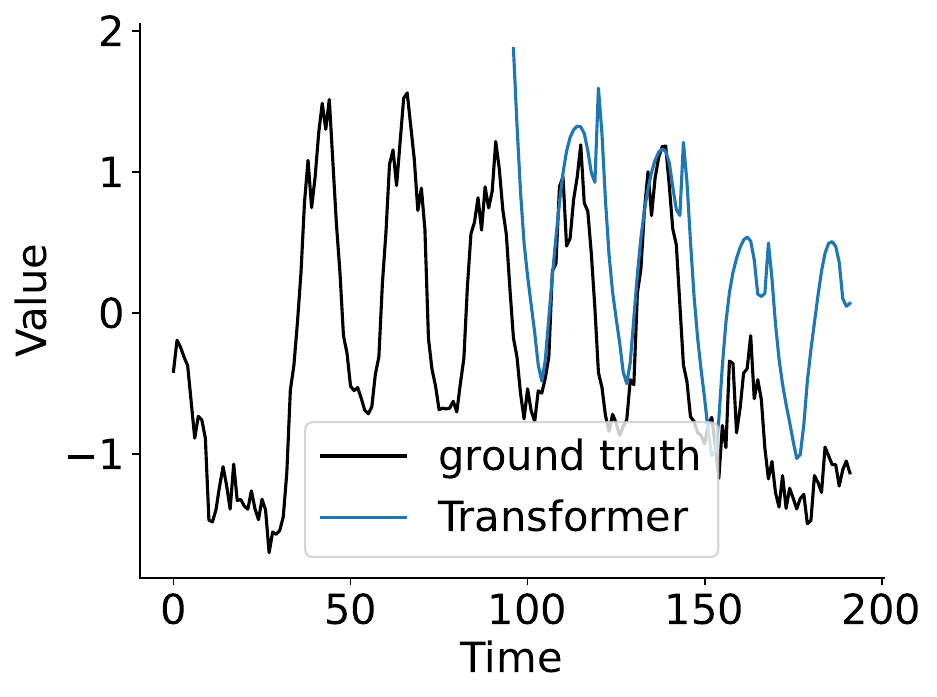}
         \caption{Comp. 3 (Side-channel)}
         \label{fig: comp 2 trans}
     \end{subfigure} \\
      \begin{subfigure}[b]{0.3\textwidth}
         \centering
         \includegraphics[width=\textwidth]{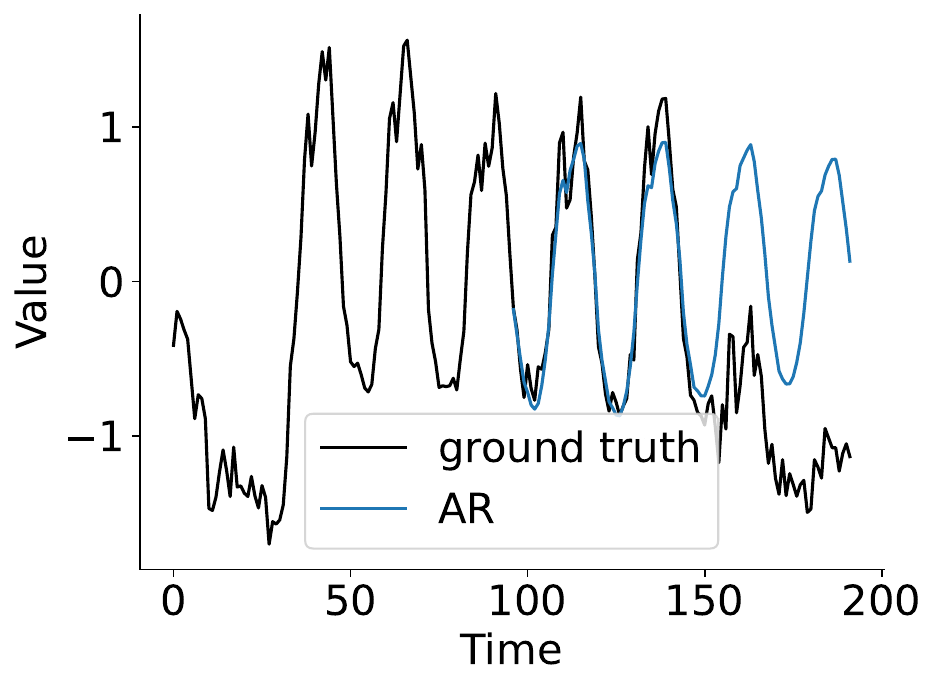}
         \caption{AR}
         \label{fig: ar app}
     \end{subfigure}
      \begin{subfigure}[b]{0.3\textwidth}
         \centering
         \includegraphics[width=\textwidth]{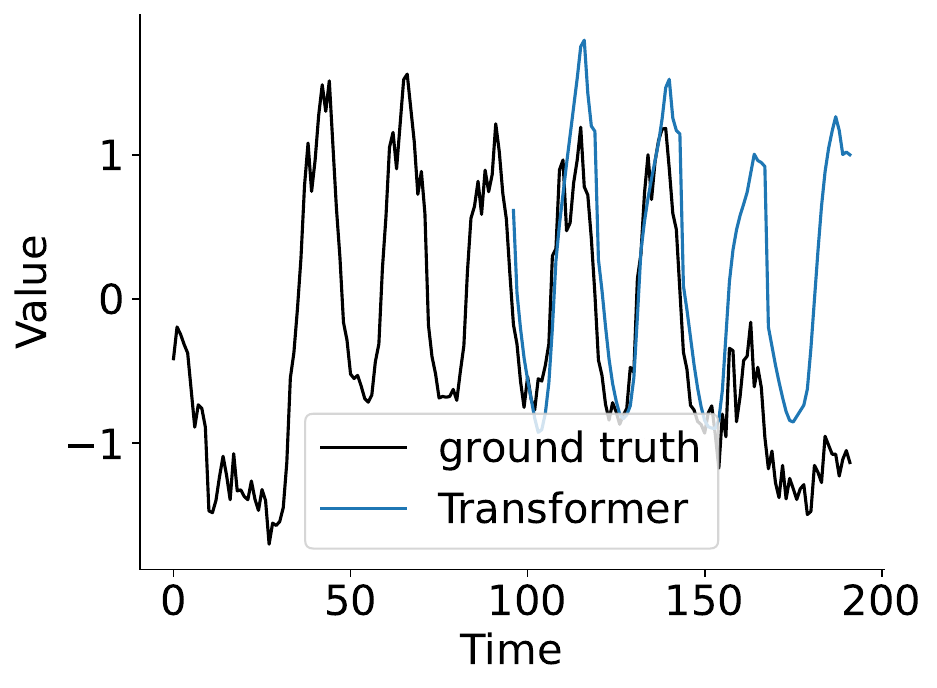}
         \caption{Full bottleneck}
         \label{fig: full bottleneck app}
     \end{subfigure}
      \begin{subfigure}[b]{0.3\textwidth}
         \centering
         \includegraphics[width=\textwidth]{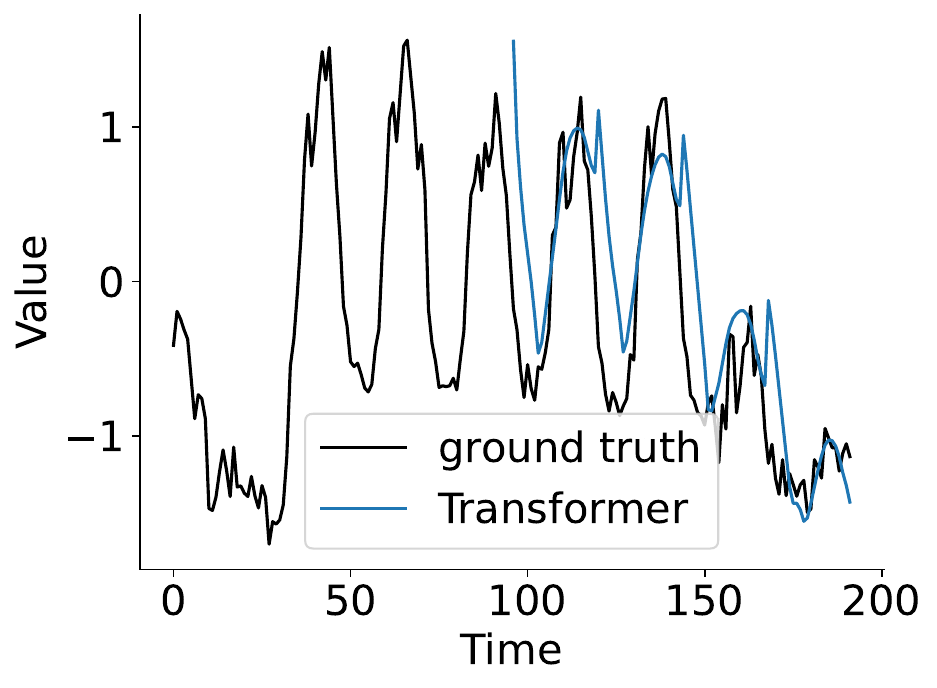}
         \caption{Final, using only comp. 3}
         \label{fig: extra comp3 final app}
     \end{subfigure}
    \caption{Vanilla Transformer forecasts from the components in the bottleneck layer (FF bottleneck on electricity data) in \ref{fig: comp 0 trans}, \ref{fig: comp 1 trans} and \ref{fig: comp 2 trans}. They are obtained by masking the other components with zero (the mean). The first half of the ground truth forms the input to the model. Note that the horizontal axes are the same across all figures, but Figure \ref{fig: comp 1 trans} contains a grid of days instead of numbered hours. Figure \ref{fig: ar app} shows the forecast made by the surrogate model AR; Figure \ref{fig: full bottleneck app} shows the forecast of the entire layer (i.e., all components together), and \ref{fig: extra comp3 final app} shows the forecast of the final layer when only the third component is used in the bottleneck layer. Note the difference between Figures \ref{fig: comp 2 trans} and \ref{fig: extra comp3 final app}, where we decode from the bottleneck and the final layer, respectively.} 
    \label{fig:comp visualizations transformer}
\end{figure}

We visualize the contributions of each component in the bottleneck using the Decoder Lens method \citep{langedijk_decoderlens_2023}, see Figure \ref{fig:comp visualizations transformer}. We obtain the output from each component individually by masking the other components with zero (close to the mean). 
Each component seems to provide similar contributions to the forecast as their respective counterpart in the Autoformer model. In particular, the first component (see Figure \ref{fig: comp 0 trans}) produces forecasts of correct seasonality and few irregularities, similar to the AR model. The second component (see Figure \ref{fig: comp 1 trans}) follows the hour-of-day feature, and the free head (see Figure \ref{fig: comp 2 trans}) picks up on high-frequency data patterns. 

\newpage
\subsection{Intervention} \label{app: vanilla intervention}
\begin{figure}[h!]
     \centering
     
     \begin{subfigure}[b]{0.4\textwidth}
         \centering
         \includegraphics[width=\textwidth]{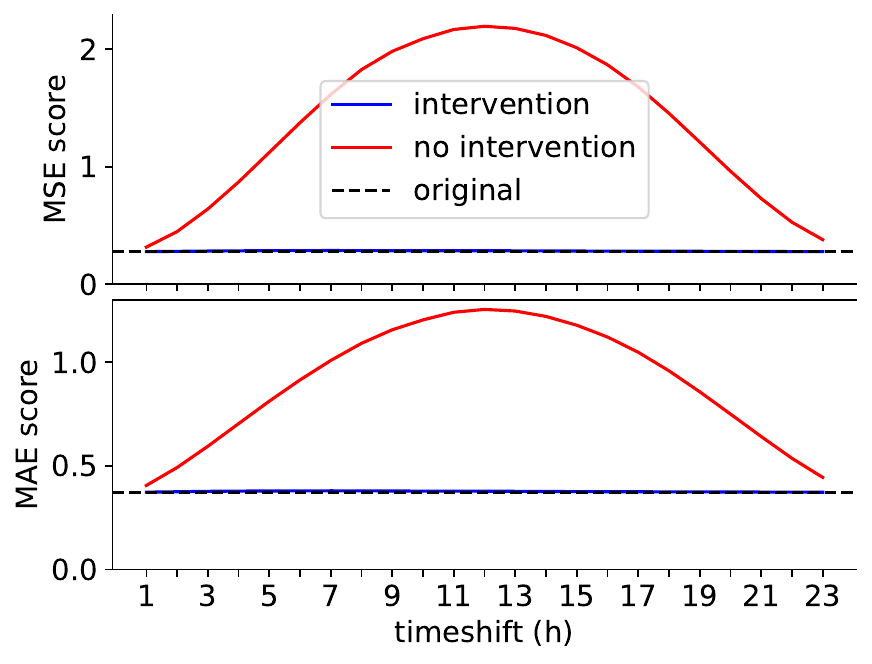}
         \caption{Attention bottleneck}
         \label{fig:att bottleneck}
     \end{subfigure}
     \begin{subfigure}[b]{0.4\textwidth}
         \centering
         \includegraphics[width=\textwidth]{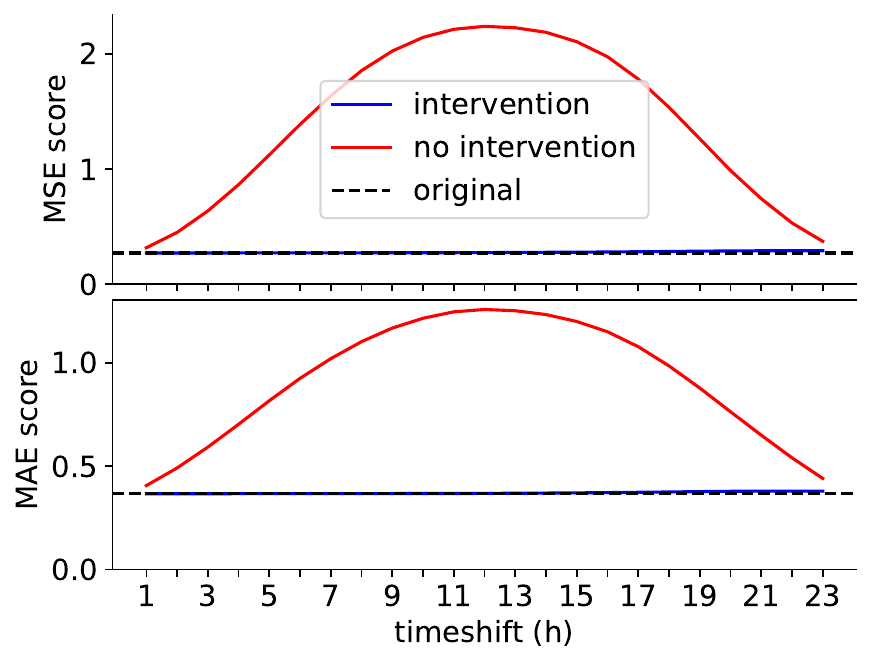}
         \caption{Feed-forward bottleneck}
         \label{fig:ff bottleneck}
     \end{subfigure}
        \caption{Performance of the bottleneck vanilla Transformer on electricity data with shifted timestamps. The dashed line represents the performance of the same model on the original data, i.e., with no timeshift.}
        \label{fig:intervention transformer}
\end{figure}

We perform the intervention experiment in the same set-up as for the Autoformer model. That is, we delay the input timestamps with a fixed number of hours to obtain shifted timestamps, and perform an intervention in the bottleneck by substituting the activations based on the shifted time with the activations from the original time. We use a vanilla Transformer trained on the electricity dataset, and perform shifts of up to and including 23 hours. We compare the performance of the intervention with out-of-the-box performance of the same model on the shifted dataset. 
The results are shown in Figure \ref{fig:intervention transformer}. For both types of bottlenecks, the intervention performs best for all timeshifts, by keeping the error scores marginally close to the original performance (with no timeshift). This indicates that the model effectively learns to represent the hour-of-day concept in the dedicated head, which is able to provide control over the model's behavior.

\subsection{Conclusion}
By repeating the set of experiments for the vanilla Transformer model, we provided further evidence for the generality of the concept bottleneck framework. In particular, we showed that the framework can be applied to the vanilla Transformer model, without having any significant impact on the overall model performance, while providing improved interpretability.

\newpage
\section{Application of Framework to FEDformer} \label{app: framework extension FEDformer}
To demonstrate the generality of our concept bottleneck framework, we apply it to \textit{FEDformer} \citep{zhou_fedformer_2022}. This is a Transformer architecture containing \textit{Fourier enhanced blocks} and \textit{wavelet enhanced blocks} to represent time series in the frequency domain. For more details, we refer to the original authors \cite{zhou_fedformer_2022}. We train the model on the same six datasets and perform an interpretability analysis.

\subsection{Performance Analysis}
The performance of the FEDformer with and without bottleneck is given in Table \ref{tab: app fedformer}. We train the bottleneck with a `free' component (the side channel), i.e., with $c=3$. Note that the model by \cite{zhou_fedformer_2022} is of a different size (two encoder layers with eight heads per layer). Interestingly, we find for some datasets (e.g. electricity and illness) that including a bottleneck increases the performance, while it has little effect on the performance for the other datasets. We can conclude for all datasets that including a bottleneck does not hurt performance.

\floatsetup[table]{objectset=centering,capposition=top}
    \begin{table}[h!]
    \captionsetup{}
        \caption{Performance of FEDformer. For both metrics, it holds that a lower score indicates a better performance, where the best results are \textbf{bold}, and the second-best are \underline{underlined}. }\label{tab: app fedformer}
        \resizebox{\textwidth}{!}{
        \begin{tabular}{lrrrrrrrr|rr}\toprule
            &\multicolumn{2}{c}{\textbf{Att bottleneck}}&\multicolumn{2}{c}{\textbf{FF bottleneck}}&\multicolumn{2}{c}{\textbf{No bottleneck}}&\multicolumn{2}{c|}{\textbf{AR}}&\multicolumn{2}{c}{\textbf{Zhou et al.}}
            \\\cmidrule(r){2-3}\cmidrule(r){4-5}\cmidrule(r){6-7}\cmidrule(r){8-9}\cmidrule(r){10-11}  
            &MSE&MAE&MSE&MAE&MSE&MAE&MSE&MAE&MSE&MAE\\\midrule

Electricity   & \textbf{0.185}       & \textbf{0.302}       & \underline{0.186}          & \underline{0.303}         & 0.189              & 0.304                 & 0.497          & 0.522          & 0.193               & 0.308              \\
Traffic       & 0.585                & \underline{0.364}          & 0.585                & \underline{0.364}         & \underline{0.573}        & \textbf{0.358}        & \textbf{0.420} & 0.494          & 0.587               & 0.366              \\
Weather       & 0.221                & 0.299                & 0.219                & 0.296               & 0.334              & 0.397                 & \textbf{0.006} & \textbf{0.062} & \underline{0.217}         & \underline{0.296}        \\
Illness       & \underline{3.070}          & \underline{1.217}          & 3.076                & 1.219               & 3.111              & 1.232                 & \textbf{1.027} & \textbf{0.820} & 3.228               & 1.260              \\
Exchange rate & 0.147                & 0.277                & \underline{0.145}          & \underline{0.275}         & 0.146              & 0.276                 & \textbf{0.082} & \textbf{0.230} & 0.148               & 0.278              \\
ETT           & 0.079                & 0.193                & 0.079                & 0.192               & \underline{0.077}        & \underline{0.190}           & \textbf{0.034} & \textbf{0.117} & 0.203               & 0.287             
            \\\bottomrule
        \end{tabular}}
    \end{table} 

\subsection{CKA Analysis}

\begin{figure}[htbp!]
\captionsetup{}
     \centering
     
     \begin{subfigure}[b]{0.35\textwidth}
         \centering \includegraphics[width=\textwidth]{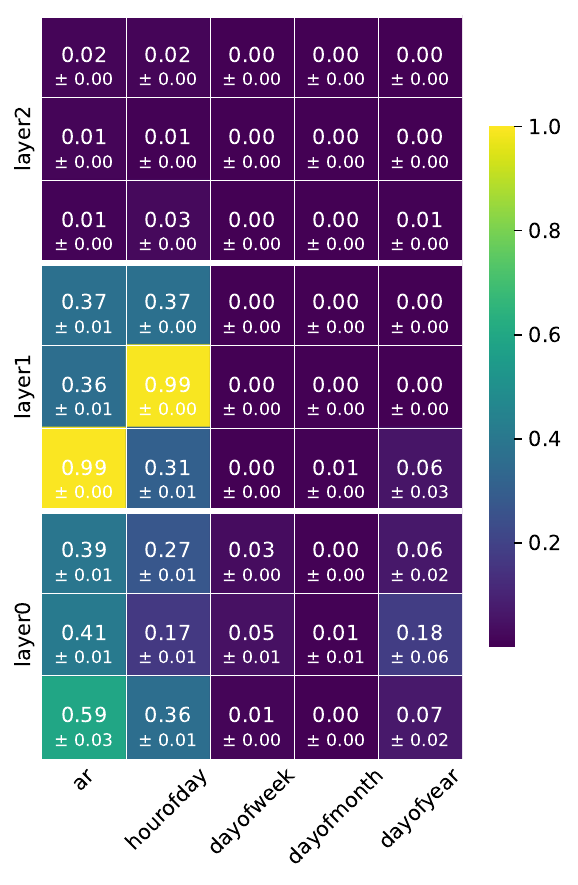}
         \caption{Att bottleneck}
         \label{}
     \end{subfigure}
     \begin{subfigure}[b]{0.35\textwidth}
         \centering
         \includegraphics[width=\textwidth]{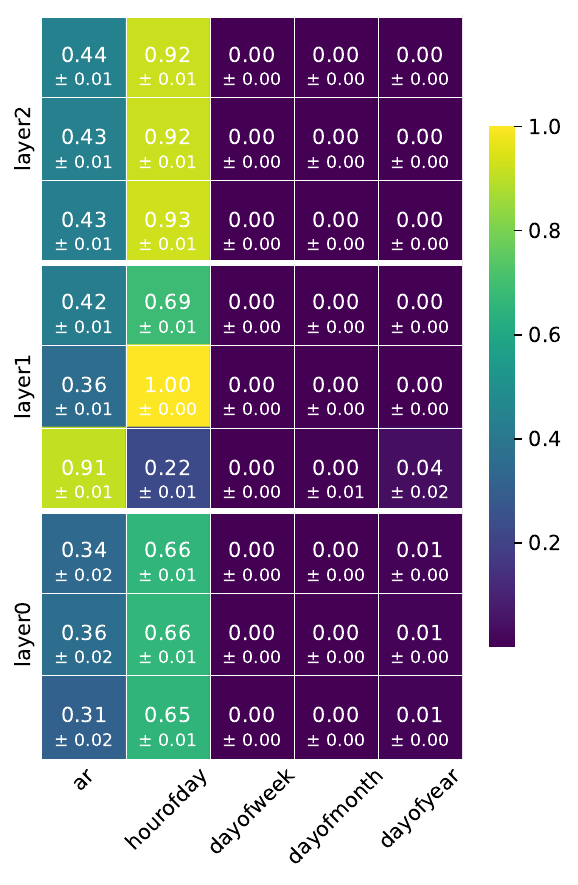}
         \caption{FF bottleneck}
         \label{}
     \end{subfigure}
    \caption{CKA scores of the FEDformer's encoder (containing three heads per layer) from the attention and feed-forward bottleneck on the electricity dataset, where each score denotes the similarity of an individual component. The first component of \texttt{layer1}  is trained to be similar to AR, and the second component to the hour-of-day concept (lower and middle row in the figure, respectively). The scores are calculated using three batches of size 32 from the test data set.}
    \label{fig:CKA scores fedformer}
\end{figure}

After training the FEDformer with our concept bottleneck framework, we evaluate the similarity of the hidden representations to the interpretable concepts using CKA, see Figure \ref{fig:CKA scores fedformer}. Recall that CKA scores are defined in the range from 0 to 1, where 1 indicates perfect similarity. Both components in the two types of bottleneck show a very high similarity to their target concept, indicating a successful training on interpretability.

\subsection{Intervention}

\begin{figure}[h!]
\captionsetup{}
     \centering
     
     \begin{subfigure}[b]{0.4\textwidth}
         \centering
         \includegraphics[width=\textwidth]{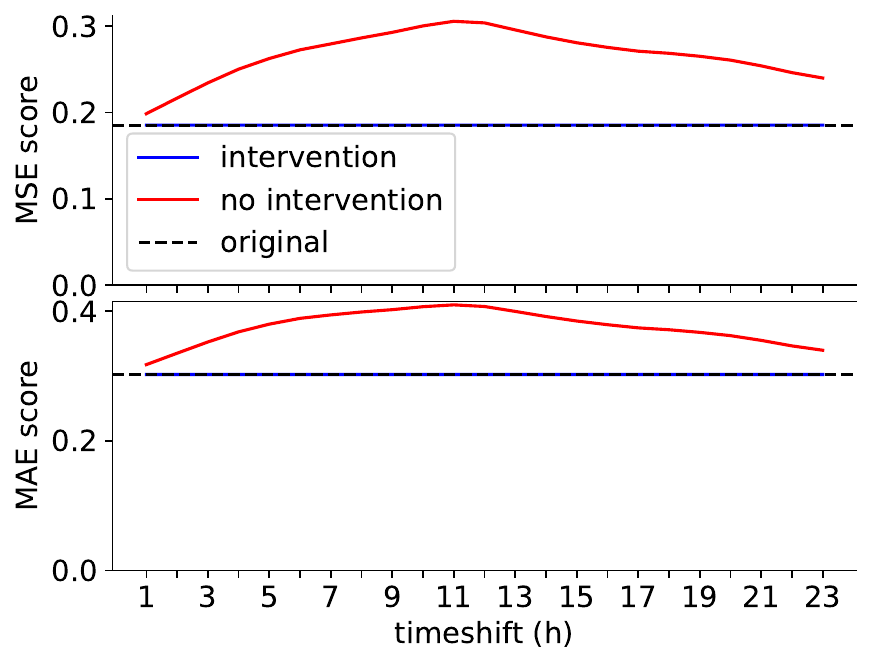}
         \caption{Attention bottleneck}
         \label{fig:att bottleneck fed}
     \end{subfigure}
     \begin{subfigure}[b]{0.4\textwidth}
         \centering
         \includegraphics[width=\textwidth]{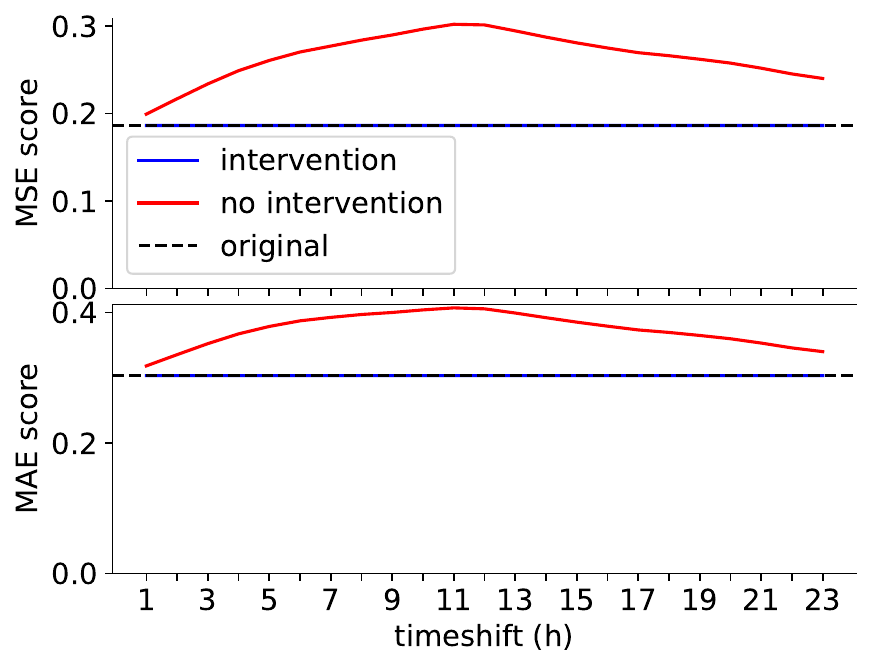}
         \caption{Feed-forward bottleneck}
         \label{fig:ff bottleneck fed}
     \end{subfigure}
        \caption{Performance of the bottleneck FEDformer on electricity data with shifted timestamps. The dashed line represents the performance of the same model on the original data, i.e., with no timeshift.}
        \label{fig:intervention fedformer}
\end{figure}

Additionally, we perform the intervention experiment in the same set-up as for the other Transformer models. That is, we delay the input timestamps with a fixed number of hours and perform an intervention in the bottleneck by substituting the activations with those based on the original time. We compare the performance of the intervention with out-of-the-box performance of the same model on the shifted dataset. The results are shown in Figure \ref{fig:intervention fedformer}. For both types of bottlenecks, the intervention performs best for all timeshifts, by keeping the error marginally close to the original performance (without timeshift). This indicates that the model effectively learns to represent the hour-of-day concept in the dedicated head, which is able to provide control over the model's behavior.

\subsection{Conclusion}
By repeating the set of experiments for the FEDformer model, we provided further evidence for the generality of the concept bottleneck framework. In particular, we showed that the framework can be applied to the FEDformer model, without having any significant impact on the overall model performance, while providing improved interpretability.

\newpage
\section{Synthetic Data} \label{app:synthetic}


To increase the understanding of how the concepts in the bottleneck can be leveraged, we train the model on a synthetic dataset. 

\subsection{Dataset}
\begin{figure}[h!]
    \centering
    \includegraphics[width=0.8\linewidth]{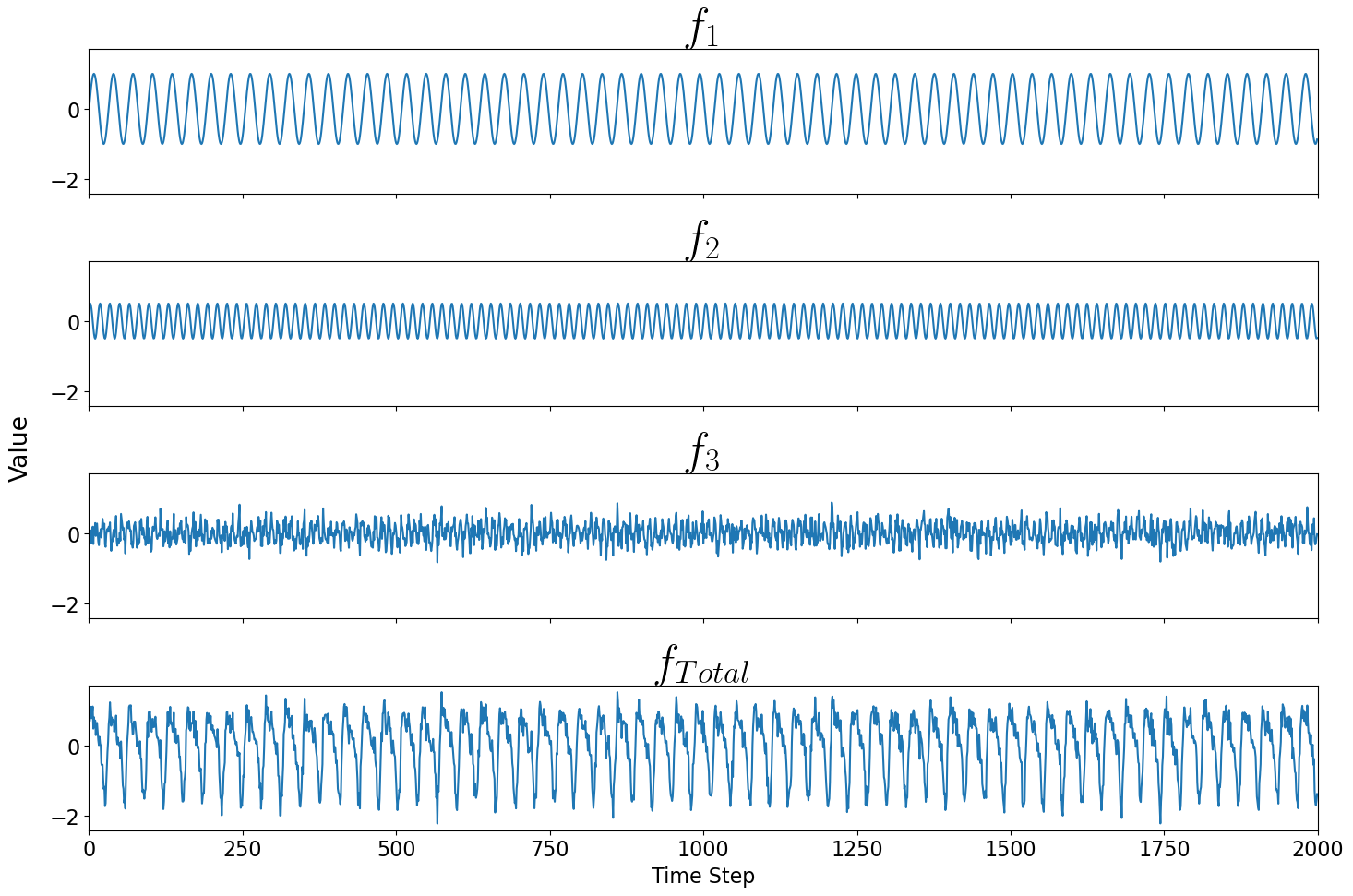}
    \caption{The synthetic time series dataset.}
    \label{fig:synthetic data}
\end{figure}

We generate a synthetic time series as the sum of different functions. In particular, the dataset is generated using the function $f_{Total}$ with time $t$ as follows:
\begin{equation*}
    f_{Total}(t) = f_1(t) + f_2(t) + f_3(t),
\end{equation*}
where:
\begin{align*}
    f_{1}(t) &= \sin(2 \pi t), \\
    f_{2}(t) &= \frac{1}{2} \sin(4 \pi t + \frac{\pi}{4}), \\
    f_{3}(t) &= \frac{1}{4} \sin(6 \pi t + \frac{\pi}{2}) + \eps_t.
\end{align*}
Note that all functions $f_1, f_2$ and $f_3$ follow a periodic structure, and $f_3$ contains random noise $\eps$ from a normal distribution with standard deviation of 0.2. See Figure \ref{fig:synthetic data} for a visualization of the functions. 

\subsection{Experiment and Results}

We train an Autoformer model on the synthetic dataset using the concept bottleneck framework. Each concept in the bottleneck is defined as one of the underlying functions (i.e., $f_1$, $f_2$ or $f_3$), for which the ground-truth is known by construction. The model contains three encoder layers, with three attention heads per layer. We apply the bottleneck to the attention heads of the second encoder layer. Additionally, we train the bottleneck using different values for hyperparameter $\alpha$, which controls the weight of the CKA loss in the total loss function (see Section \ref{sec:loss}). 

\begin{figure}[htbp!]
     \centering
     \includegraphics[width=0.5\textwidth]{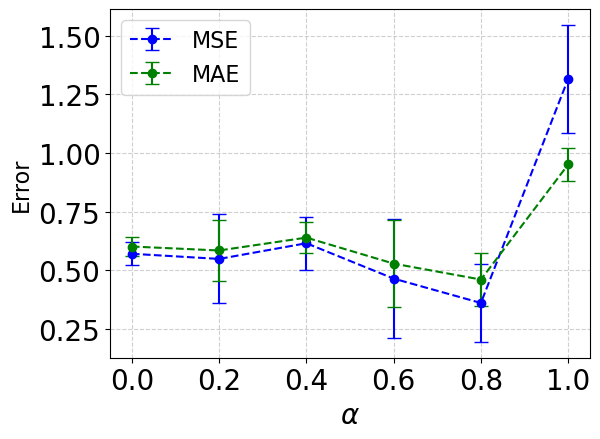}
     \caption{Performance on the synthetic dataset for different values of $\alpha$, using an Autoformer with attention bottleneck. For both metrics, it holds that a lower score indicates a better performance. The standard deviation is provided over 5 different seeds.}
     \label{fig:app alphas synthetic}
\end{figure}

\begin{figure}[h!]
     \centering

     
     \begin{subfigure}[b]{0.45\textwidth}
         \centering
         \includegraphics[width=\textwidth]{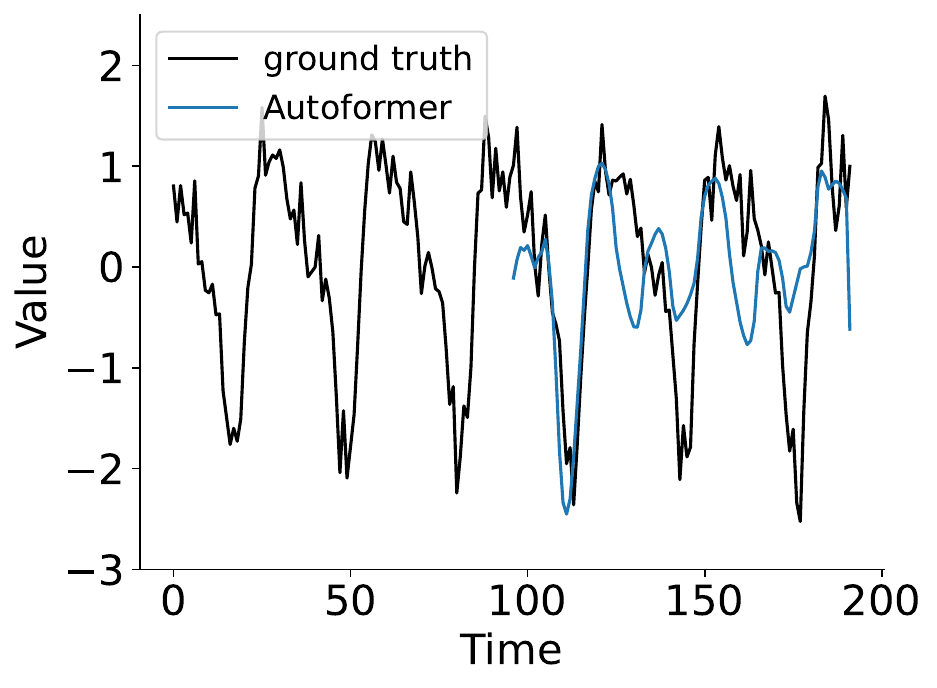}
         \caption{$\alpha=0.0$}
         \label{}
     \end{subfigure}
     \begin{subfigure}[b]{0.45\textwidth}
         \centering
         \includegraphics[width=\textwidth]{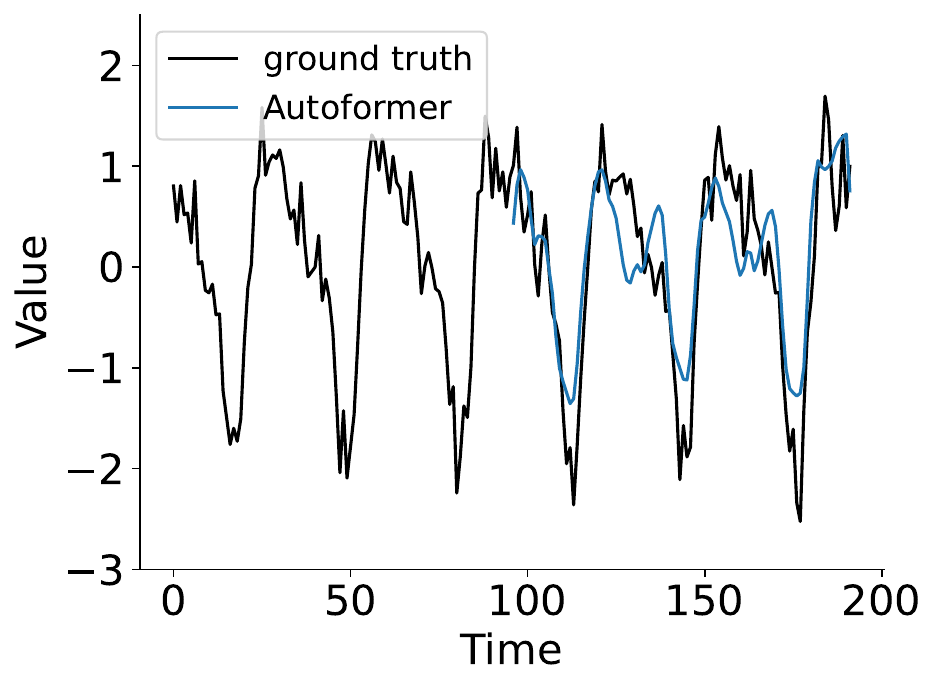}
         \caption{$\alpha=0.4$}
         \label{}
     \end{subfigure}
     \begin{subfigure}[b]{0.45\textwidth}
         \centering
         \includegraphics[width=\textwidth]{images/synthetic/08_final_all.pdf}
         \caption{$\alpha=0.8$}
         \label{}
     \end{subfigure}
     \begin{subfigure}[b]{0.45\textwidth}
         \centering
         \includegraphics[width=\textwidth]{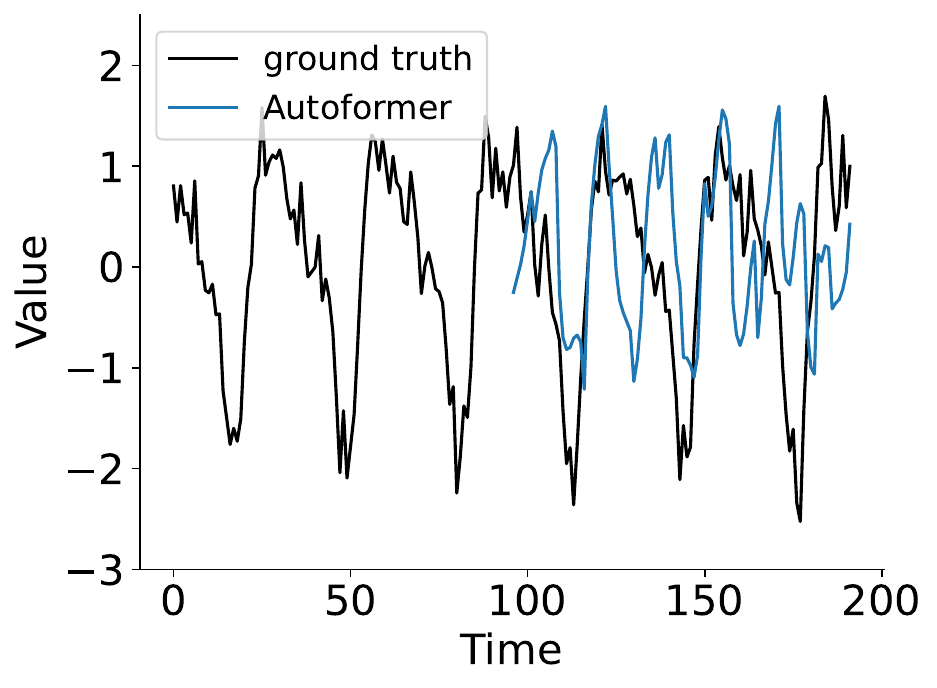}
         \caption{$\alpha=1.0$}
         \label{}
     \end{subfigure}
        \caption{Predictions of the Autoformer model on a sample from the test dataset. The Autoformer is trained with an attention bottleneck using different values of hyperparameter $\alpha$ and the same seed.}
        \label{fig:plots predictions synthetic}
\end{figure}

As expected, we find for all values $\alpha < 1$ that the model is able to forecast the dataset well, see Figure \ref{fig:app alphas synthetic}. Note that a low forecasting error cannot be expected for $\alpha=1$, because in this edge case the loss function does not contain any term that represents the forecasting error. Remarkably, for all other cases, the performance of the Autoformer seems to improve as $\alpha$ increases. This suggests that properly chosen concepts improve the performance of the model, at least when the ground-truth underlying functions are known. It should be noted that the standard deviation is higher for all $\alpha > 0$, which indicates that initialization of the parameters is important when learning the bottleneck. Additionally, visualizations of the predictions are given in Figure \ref{fig:plots predictions synthetic}.



\begin{figure}[htbp!]
     \centering
     
     \begin{subfigure}[b]{0.32\textwidth}
         \centering
         \includegraphics[width=\textwidth]{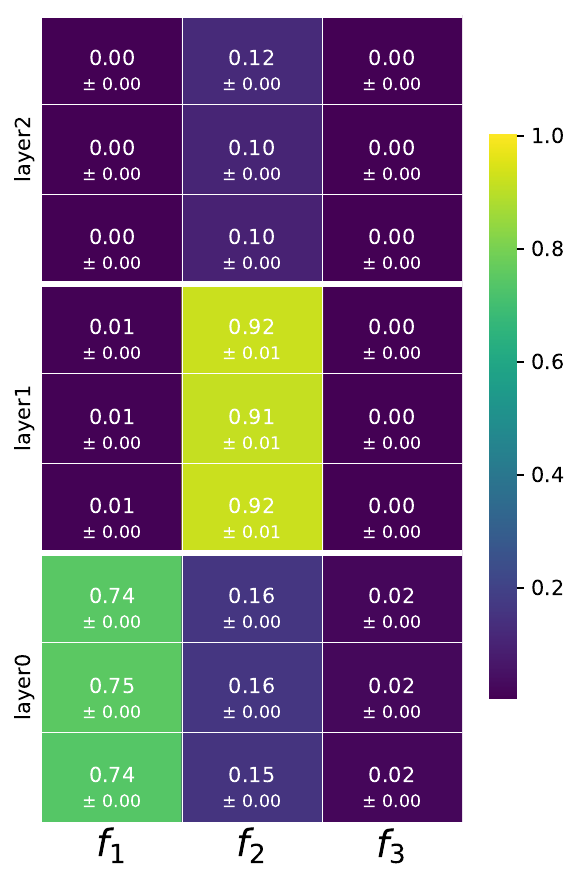}
         \caption{$\alpha=0$}
         \label{fig:alpha is 0}
     \end{subfigure}
     \begin{subfigure}[b]{0.32\textwidth}
         \centering
         \includegraphics[width=\textwidth]{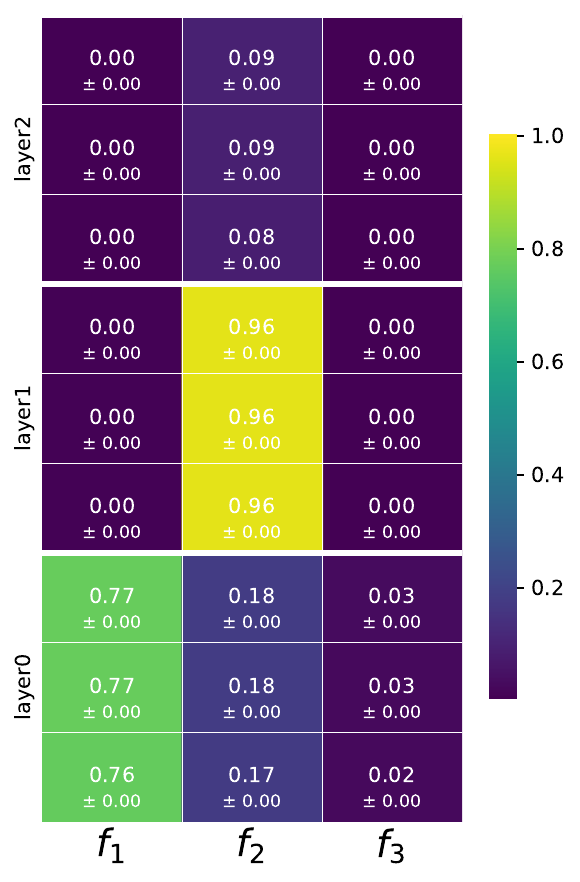}
         \caption{$\alpha=0.2$}
         \label{}
     \end{subfigure}
     \begin{subfigure}[b]{0.32\textwidth}
         \centering
         \includegraphics[width=\textwidth]{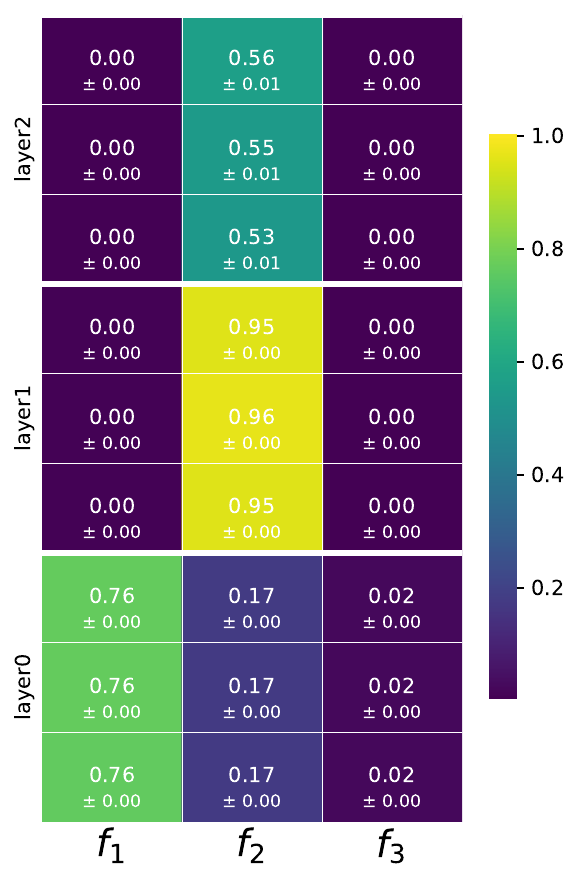}
         \caption{$\alpha=0.4$}
         \label{}
     \end{subfigure}
     \centering
     \begin{subfigure}[b]{0.32\textwidth}
         \centering
         \includegraphics[width=\textwidth]{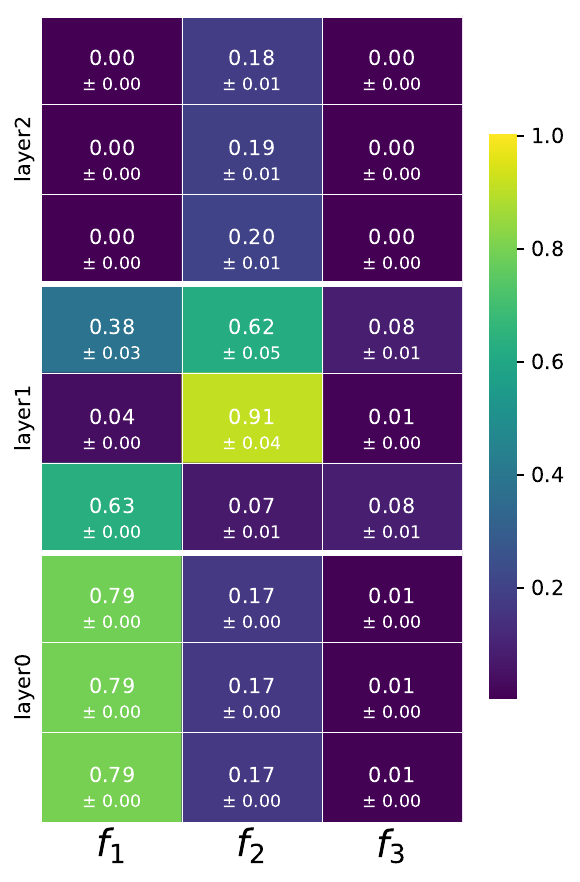}
         \caption{$\alpha=0.6$}
         \label{fig:alpha is 0}
     \end{subfigure}
     \begin{subfigure}[b]{0.32\textwidth}
         \centering
         \includegraphics[width=\textwidth]{images/synthetic/CKA_plot_parallel_08.pdf}
         \caption{$\alpha=0.8$}
         \label{}
     \end{subfigure}
     \begin{subfigure}[b]{0.32\textwidth}
         \centering
         \includegraphics[width=\textwidth]{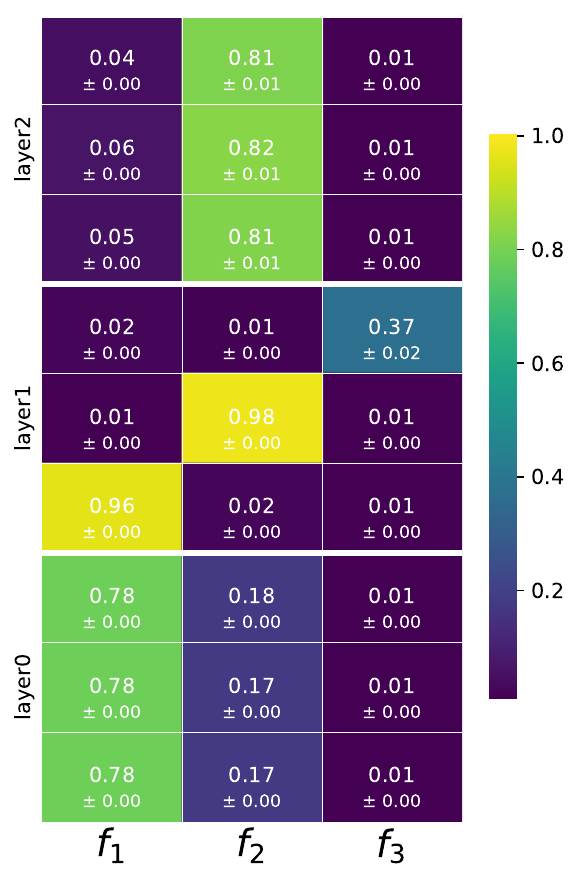}
         \caption{$\alpha=1$}
         \label{}
     \end{subfigure}
    \caption{CKA scores of the attention bottleneck Autoformer on synthetic data for different values of hyperparameter $\alpha$. The scores are calculated using three batches of size 32 of the test data set.}
    \label{fig:app alpha synthetic}
\end{figure}

Additionally, the different values of hyperparameter $\alpha$ show clearly how the different concepts are leveraged by the model, see Figure \ref{fig:app alpha synthetic}. The figure shows the similarity scores between the attention heads and the different underlying functions of the dataset. Without the CKA loss, at $\alpha=0$, the different heads in \texttt{layer1} of the model do not show high similarity to their respective concepts, i.e., functions. Instead, all heads have a high similarity to concept $f_2$. This is different for higher values of $\alpha$, where the different heads show higher similarity to their respective concepts. Note that the third concept $f_3$ cannot be perfectly learned by the model because of the random noise component. 

All in all, these results show that a higher value for $\alpha$, which is equivalent to a higher weight of the CKA loss in the total loss function, results in more similarity of the bottleneck components to their respective concepts, as expected.

\newpage
\section{Effect of AR as Surrogate Model} \label{app:ar}

Interestingly, the AR model outperforms the Autoformer for some datasets (see Table \ref{tab:performance analysis}). This raises the question whether the AR surrogate model makes up for any loss in performance introduced by the concept bottleneck.  

To test this, we train an Autoformer without the AR concept. Specifically, we include the time concept and a free component in the feed-forward bottleneck. Here, the free component refers to a component in the bottleneck that is not included in the CKA loss (see Section \ref{sect: bottleneck}). 

\begin{figure}[htbp!]
     \centering
     
     \begin{subfigure}[b]{0.45\textwidth}
         \centering
         \includegraphics[width=\textwidth]{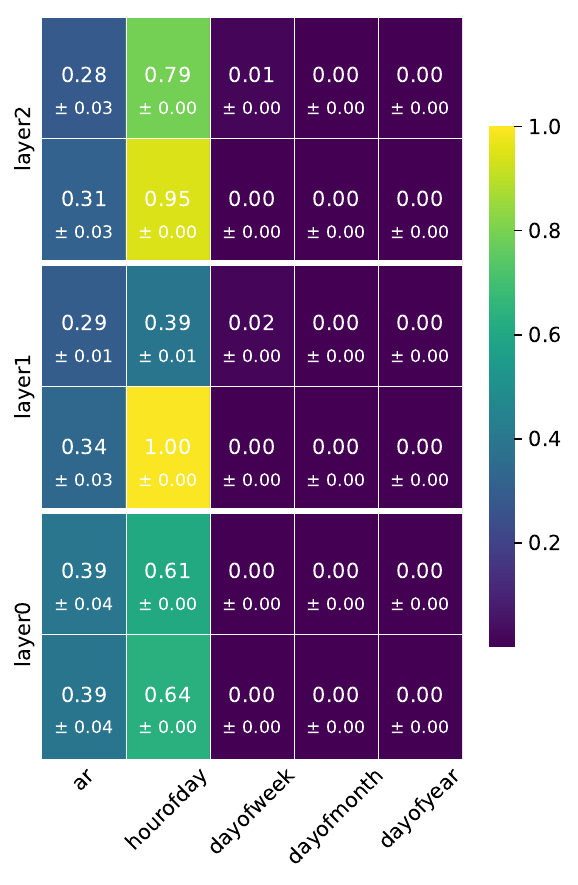}
         \caption{Without AR (MSE: 0.206, MAE: 0.321)}
         \label{fig:without ar}
     \end{subfigure}
     \begin{subfigure}[b]{0.45\textwidth}
         \centering
         \includegraphics[width=\textwidth]{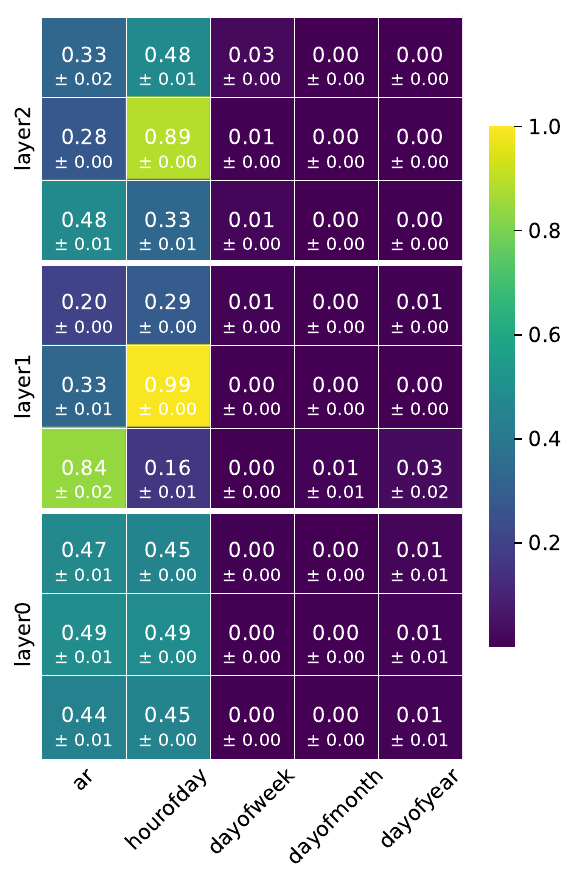}
         \caption{With AR (MSE: 0.207, MAE: 0.320)}
         \label{fig:with ar}
     \end{subfigure}
    \caption{CKA plots of two Autoformer models with feed-forward bottlenecks. The model in \ref{fig:without ar} is trained without AR in the bottleneck, while the model in \ref{fig:with ar} is trained with AR. Note that the upper component in \texttt{layer1} is the free component in both plots.}
    \label{fig:cka_plots_effect_ar}
\end{figure}

The performance on the electricity data for this model is (MSE: 0.206, MAE: 0.321), which is seemingly identical to the original performance of (MSE: 0.207, MAE: 0.320). This suggests that it is not the AR head that makes up for the loss in performance. The CKA plots, see Figure \ref{fig:cka_plots_effect_ar}, verify that there is no component in the minimal set-up (without AR) that is very similar to the AR model, unlike in the original set-up. So, these results show that the AR model does not add performance to the bottleneck model, merely  interpretability.

Additionally, we refer the reader to Appendix \ref{app:synthetic}, where we perform more experiments on training the bottleneck without the AR surrogate model.


\newpage
\section*{NeurIPS Paper Checklist}

\begin{enumerate}

\item {\bf Claims}
    \item[] Question: Do the main claims made in the abstract and introduction accurately reflect the paper's contributions and scope?
    \item[] Answer: \answerYes{} 
    \item[] Justification: The claims match theoretical and experimental results, and reflect how much the results can be expected to generalize to other settings. 

\item {\bf Limitations}
    \item[] Question: Does the paper discuss the limitations of the work performed by the authors?
    \item[] Answer: \answerYes{} 
    \item[] Justification: Section \ref{sect: discussion} discusses the limitations of the work, including the assumptions and computational efficiency of the method.

\item {\bf Theory assumptions and proofs}
    \item[] Question: For each theoretical result, does the paper provide the full set of assumptions and a complete (and correct) proof?
    \item[] Answer: \answerNA{} 
    \item[] Justification: The paper does not include theoretical results.

    \item {\bf Experimental result reproducibility}
    \item[] Question: Does the paper fully disclose all the information needed to reproduce the main experimental results of the paper to the extent that it affects the main claims and/or conclusions of the paper (regardless of whether the code and data are provided or not)?
    \item[] Answer: \answerYes{} 
    \item[] Justification: Section \ref{sect: method} discloses all information needed to reproduce the main experimental results. Additionally, Appendix \ref{app: formalization} provides a formalization of the proposed method. 

\item {\bf Open access to data and code}
    \item[] Question: Does the paper provide open access to the data and code, with sufficient instructions to faithfully reproduce the main experimental results, as described in supplemental material?
    \item[] Answer: \answerNo{} 
    \item[] Justification: Upon acceptance of the paper in an archive, we will release the code. The data is open-source, and already available.

\item {\bf Experimental setting/details}
    \item[] Question: Does the paper specify all the training and test details (e.g., data splits, hyperparameters, how they were chosen, type of optimizer, etc.) necessary to understand the results?
    \item[] Answer: \answerYes{} 
    \item[] Justification: Section \ref{sect: implementation} specificies the implementation details.

\item {\bf Experiment statistical significance}
    \item[] Question: Does the paper report error bars suitably and correctly defined or other appropriate information about the statistical significance of the experiments?
    \item[] Answer: \answerYes{} 
    \item[] Justification: Appendix \ref{app: detailed results} and \ref{app: sensitivity} report error margins and details on hyperparameter sensitivity.

\item {\bf Experiments compute resources}
    \item[] Question: For each experiment, does the paper provide sufficient information on the computer resources (type of compute workers, memory, time of execution) needed to reproduce the experiments?
    \item[] Answer: \answerYes{} 
    \item[] Justification: Section \ref{sect: implementation} provides the information on computer resources.
    
\item {\bf Code of ethics}
    \item[] Question: Does the research conducted in the paper conform, in every respect, with the NeurIPS Code of Ethics \url{https://neurips.cc/public/EthicsGuidelines}?
    \item[] Answer: \answerYes{} 
    \item[] Justification: There are no violations with the NeurIPS Code of Ethics.

\item {\bf Broader impacts}
    \item[] Question: Does the paper discuss both potential positive societal impacts and negative societal impacts of the work performed?
    \item[] Answer: \answerYes{} 
    \item[] Justification: Potential social impacts are discussed in Section \ref{sect: discussion}.
    
\item {\bf Safeguards}
    \item[] Question: Does the paper describe safeguards that have been put in place for responsible release of data or models that have a high risk for misuse (e.g., pretrained language models, image generators, or scraped datasets)?
    \item[] Answer: \answerNA{} 
    \item[] Justification: The paper poses no such risks.

\item {\bf Licenses for existing assets}
    \item[] Question: Are the creators or original owners of assets (e.g., code, data, models), used in the paper, properly credited and are the license and terms of use explicitly mentioned and properly respected?
    \item[] Answer: \answerYes{} 
    \item[] Justification: The creators of the assets used in the paper are cited.

\item {\bf New assets}
    \item[] Question: Are new assets introduced in the paper well documented and is the documentation provided alongside the assets?
    \item[] Answer: \answerNA{} 
    \item[] Justification: The paper does not release new assets.

\item {\bf Crowdsourcing and research with human subjects}
    \item[] Question: For crowdsourcing experiments and research with human subjects, does the paper include the full text of instructions given to participants and screenshots, if applicable, as well as details about compensation (if any)? 
    \item[] Answer: \answerNA{} 
    \item[] Justification: The paper does not involve crowdsourcing nor research with human subjects.

\item {\bf Institutional review board (IRB) approvals or equivalent for research with human subjects}
    \item[] Question: Does the paper describe potential risks incurred by study participants, whether such risks were disclosed to the subjects, and whether Institutional Review Board (IRB) approvals (or an equivalent approval/review based on the requirements of your country or institution) were obtained?
    \item[] Answer: \answerNA{} 
    \item[] Justification: The paper does not involve crowdsourcing nor research with human subjects.

\item {\bf Declaration of LLM usage}
    \item[] Question: Does the paper describe the usage of LLMs if it is an important, original, or non-standard component of the core methods in this research? Note that if the LLM is used only for writing, editing, or formatting purposes and does not impact the core methodology, scientific rigorousness, or originality of the research, declaration is not required.
    \item[] Answer: \answerNA{} 
    \item[] Justification: The core method development in this research does not involve LLMs as any important, original, or non-standard components.

\end{enumerate}

\end{document}